\documentclass{article}

\usepackage{microtype}
\usepackage{graphicx}
\usepackage{subcaption}
\usepackage{booktabs} 

\usepackage{hyperref}

\usepackage[preprint]{icml2026}

\usepackage{amsmath}
\usepackage{amssymb}
\usepackage{mathtools}
\usepackage{amsthm}
\usepackage{natbib}
\usepackage{enumitem}
\usepackage{comment}
\usepackage{float}
\usepackage{makecell}

\usepackage[capitalize,noabbrev]{cleveref}

\theoremstyle{plain}
\newtheorem{theorem}{Theorem}[section]

\theoremstyle{definition}
\newtheorem{definition}[theorem]{Definition}

\theoremstyle{remark}
\newtheorem{remark}[theorem]{Remark}

\usepackage[textsize=tiny, disable]{todonotes}

\icmltitlerunning{When, Where, How: AICL4TFMs}

\begin{document}

\twocolumn[


\icmltitle{When, Where, How: Active In-Context Learning for Tabular Foundation Models}



  \icmlsetsymbol{equal}{*}

  \begin{icmlauthorlist}
    \icmlauthor{Christian M. M. Frey}{utn}
  \end{icmlauthorlist}
  \icmlaffiliation{utn}{Department of Computer Science \& Artificial Intelligence, University of Technology Nuremberg, Dr.-Luise-Herzberg-Straße 4, 90461 Nuremberg, Germany}

  \icmlcorrespondingauthor{Christian Frey}{Christian.Frey@utn.de}

  \icmlkeywords{Machine Learning, ICML}

  \vskip 0.3in
]




\printAffiliationsAndNotice{}  

\begin{abstract}
    Tabular Foundation Models (TFMs) adapt to a new task only through the labeled rows in their context.
    However, labels are expensive to acquire in real-world scenarios due to the need for highly specific domain knowledge to ensure high-quality data. Active in-context learning transfers classical acquisition rules to this setting, in which the model that guides acquisition changes with every label added. 
    We ask how the value of diversity, predictive uncertainty and target-aware information gain changes as a frozen TFM acquires more context. For that, we study \textit{\textbf{T}arget-\textbf{A}ware \textbf{N}earest \textbf{N}eighbor} (\textbf{TANN}) acquisition and the \textit{\textbf{E}xpected \textbf{P}redictive \textbf{I}nformation Gain in \textbf{C}ontext}~(\textbf{EPIC}) and evaluate them against classical rules. 
    Our study across twenty benchmarks with two well-established TFMs and classical GBDTs highlights three findings: (i)~\textit{When:} diversity-based acquisition is best during the first acquisition iterations; (ii)~\textit{Where:} labeling pool rows that serve the target inputs via a simple target-aware nearest-neighbor rule matters substantially; (iii)~\textit{How:} EPIC yields the best models at the label budget, retains on average 80--87\% of the chance-corrected full-data performance with only 5--10\% of the pool labeled on the eight benchmarks with the largest pools, and selects labels that also improve the predictive performance of gradient-boosted decision trees.
\end{abstract}

\section{Introduction}
\label{sec:intro}

\todo{good text outcommented}

Tabular foundation models (TFMs) like TabPFN~\citep{hollmann2023tabpfn,
hollmann2025tabpfn} and TabICL~\citep{qu2025tabicl,qu2026tabiclv2}, pre-trained on a large corpus of synthetic tasks, solve new tasks by in-context learning \citep{dong_2024_iclsurvey}. The frozen models are given labeled rows together with the query rows and return predictions in a single forward pass. The labeled rows are therefore the only information that adapts a TFM to a new task, and they have a decisive influence on its predictive performance. This, however, presupposes access to labeled data that can be fed to the model. In many real-world applications, labels are the bottleneck~\citep{Chai2022Selective,Tharwat2023SurveyAL,Li2023Label-Efficient}: they are the costly resource, as they require expert annotation, measurement or time, while unlabeled data are abundant. The choice of the rows to label is therefore a central lever and a question in its own right.

Recently, active in-context learning (AICL) has emerged as a paradigm to dynamically grow the context of a frozen TFM from an unlabeled pool.
Tab-AICL~\citep{treerath2026tabaicl} transfers conventional acquisition like margin-, core-set- and hybrid-based
rules~\citep{settles2009active,sener2018coreset} from classical active learning to TFMs. 
A key finding is that no rule dominates across datasets and that the gains are concentrated in the cold start. 
Their analysis also makes the assumption that the rules applied remain unchanged throughout acquisition, yet a frozen TFM changes with every acquired label. 
The quantities that acquisition rules rely on, the model's uncertainty and its view of which rows are similar, change accordingly as more labels are acquired. 
Moreover, existing rules score a candidate by its own uncertainty or by its distance to the labeled set, and ignore where the model will be used. In classical active learning, the dependence of the best strategy on the budget has been documented~\citep{hacohen2022budget}, and prediction-oriented criteria have been proposed that target the inputs of interest~\citep{yu2006transductive,
bickfordsmith2023epig}; for frozen TFMs, neither aspect has been examined.

We therefore ask: \emph{how does the value of diversity, predictive uncertainty and target-aware information gain change as a frozen tabular foundation model acquires more context?} Answering this question requires measuring the value of a label for the predictions that are actually required. For conventional active learning approaches, this value, the expected predictive information gain (EPIG)~\citep{bickfordsmith2023epig}, requires a posterior over parameters and refitting on hypothetical data. A TFM removes both obstacles: its prediction already approximates a posterior predictive distribution, and conditioning it on a hypothetical label amounts to appending one row to its context. 
We introduce the expected predictive information gain in context (EPIC), which is assessed on the target inputs through the model's in-context conditioning and evaluates all candidate-label hypotheses of a round with a single fit. Moreover, we define a target-aware nearest neighbor (TANN) approach, a simple rule that labels, for the targets on which the model is least certain, their nearest pool rows. Together with diversity- and uncertainty-based rules from prior works, these instruments separate three aspects of acquisition.

Our evaluation on the twenty benchmarks of~\citet{treerath2026tabaicl} supports three findings: 
\begin{itemize}[leftmargin=*]
  \item \textbf{When: acquisition should change as the context grows.} Diversity-based
    acquisition is the best use of the first few dozen labels, while prediction-oriented
    rules overtake it once the model's predictions become informative. 
    The pattern holds for both TabICL and TabPFN (\cref{appdx:tabpfn}), 
    where the diversity-based rule leads with 1\% of the pool labeled and prediction-oriented acquisition from 5\% on (\cref{sec:eval-when}).
  \item \textbf{Where: target awareness matters.}
    Controls that use the target inputs without uncertainty already improve on random labeling, and choosing which targets to serve improves further. The simple TANN rule is statistically
    indistinguishable from EPIC at about an eighth of its cost (\cref{sec:eval-where}). 
  \item \textbf{How: in-context conditioning makes prediction-oriented information gain computable.} EPIC implements EPIG for a frozen TFM without parameter posteriors or retraining (\cref{sec:method}). It yields the best models at the label budget, improving on the best previous rule on 15 of 20 datasets and on TabPFN-, gradient-boosting- and label-spreading-based baselines. 
    With 5\% and 10\% of the pool labeled, it retains on average 80\% and 87\% of the full-data Cohen's $\kappa$ on the eight benchmarks with the largest pools (\cref{fig:reduction}). Moreover, the labels it selects also improve classical GBDTs, XGBoost and CatBoost, over their own acquisition rules
    (\cref{sec:eval-how}).
\end{itemize}


\section{Related Work}
\label{sec:related}

\textbf{Tabular foundation models.}
TabPFN~\citep{hollmann2023tabpfn} established 
prior-data fitted networks for small classification problems, and its successor~\citep{hollmann2025tabpfn} extended it to regression, larger tables and heterogeneous features. 
TabICL~\citep{qu2025tabicl} introduced a three-stage architecture of column-wise embedding, row-wise interaction and dataset-wise in-context learning that scales in-context learning to substantially larger tables, and TabICLv2~\citep{qu2026tabiclv2} further improved its accuracy and efficiency. 
Common to these models is that labeled data enter only through the context, which has two consequences that our work builds on: adapting the model to new labels requires \textit{no training}, and the model's response to a hypothetical label can be obtained by a forward pass.

\textbf{Active learning.}
When labels are expensive, active learning chooses which instances to label~\citep{settles2009active}. Uncertainty sampling queries the instances on which the current model is least confident, for example by the margin between the two most probable classes or by predictive entropy, and diversity-based rules such as core-set selection~\citep{sener2018coreset} cover the input space with a small set of representatives; hybrid rules combine both. These criteria are proxies: they describe the candidate, not the effect its label would have on the model, and they do not depend on where predictions will be required. 
Transductive experimental design~\citep{yu2006transductive} addresses the latter by choosing the instances that best represent the unlabeled data on which predictions are needed. Classical active learning further retrains the model after every batch, which is costly and, with few labels, prone to an unstable cold start.

\textbf{Acquisition depends on the budget.}
Which of these criteria works best depends on how many labels are available. \citet{hacohen2022budget} show that representative, typical examples are most useful when labels are scarce, whereas uncertain, atypical examples become more useful as the budget grows, and propose a typicality-based rule for the low-budget regime; the cold start of uncertainty sampling, whose estimates are unreliable while the model is poorly trained, is a recurring theme of the same literature~\citep{settles2009active}. 
For a frozen TFM the learner is \textit{never trained}, and what changes with the budget is only the context it
conditions on. The context's value of diversity and uncertainty w.r.t. its size has not been studied in this regime. 

\textbf{Target-aware and information-based acquisition.}
A principled alternative measures the value of a label by the information it provides. \citet{lindley1956measure} defined the information of an experiment as the expected reduction of uncertainty about the quantity of interest, and \citet{mackay1992information} derived information-based objectives for active data selection, distinguishing information about the model parameters from information about predictions at chosen inputs. BALD~\citep{houlsby2011bald} selects by the mutual information between a label and the model parameters. 
Prediction-oriented Bayesian active learning~\citep{bickfordsmith2023epig} argues that parameter information can be spent on aspects of the model that are irrelevant to the predictions of interest and introduces the expected predictive information gain (EPIG), the mutual information between the label of a candidate and the label of a target input drawn from the deployment distribution. 

\textbf{Context restriction and context selection.}
Attention across context rows grows quadratically with the context size, and performance can degrade when the context leaves the range of the pre-training prior~\citep{chen2026vipcop}. 
Retrieval conditions each query on its nearest training rows: LoCalPFN~\citep{thomas2024localpfn} retrieves neighbors for TabPFN and fine-tunes the model on local contexts, and Localized TabICLv2~\citep{guta2026localized} retrieves neighbors in TabICLv2's row-representation space and retains most of the full-context accuracy at a fraction of the inference cost. Compression replaces the training set by a short learned context, either synthetic rows obtained by in-context data distillation~\citep{ma2024icd} or learned context tokens obtained by prompt tuning~\citep{feuer2024tunetables}. 
VIP-COp~\citep{chen2026vipcop} selects a subset of real rows by estimating the value of each row from the validation performance of random
contexts. 
All of these methods select the labeled rows the model sees, but not which ones to label.

\textbf{Active in-context learning and our position.}
Active in-context learning combines a frozen model, as in context selection, with label acquisition under a budget, as in active learning. Tab-AICL~\citep{treerath2026tabaicl}
formalizes this setting for TabPFN, grows the context from one row per class in batches, and adapts margin, core-set, hybrid and proxy-filtered rules; the gains over random sampling are concentrated in the cold-start regime, and no rule dominates across datasets. Each rule is applied unchanged throughout acquisition, and none uses the
inputs on which the model will be deployed. Our work addresses both aspects. 

We study how the value of diversity, uncertainty and target-aware information changes as the context grows, extending the budget dependence observed by~\citet{hacohen2022budget} to frozen in-context learners. 
We show that target awareness, as in transductive experimental design~\citep{yu2006transductive}, is a central lever in this setting, already in the form of a simple nearest-neighbor rule.
And we make prediction-oriented information gain~\citep{bickfordsmith2023epig} computable for frozen TFMs by replacing the parameter posterior with in-context
conditioning, which requires neither posterior samples nor retraining. 
To our knowledge, this is the first study of acquisition dynamics and target-aware acquisition for frozen tabular foundation models.
%

\providecommand{\R}{\mathbb{R}}
\providecommand{\Pool}{\mathcal{U}}
\providecommand{\Lab}{\mathcal{L}}
\providecommand{\Tgt}{\mathcal{T}}
\providecommand{\Data}{\mathcal{D}}
\providecommand{\Cand}{\mathcal{C}}
\providecommand{\Y}{\mathcal{Y}}
\providecommand{\Ent}{\mathrm{H}}
\providecommand{\MI}{\mathrm{I}}
\providecommand{\E}{\mathbb{E}}
\providecommand{\argmax}{\operatorname*{arg\,max}}
\providecommand{\argmin}{\operatorname*{arg\,min}}

\section{Method}
\label{sec:method}

\begin{figure*}[t]
    \centering
    \includegraphics[width=1\linewidth]{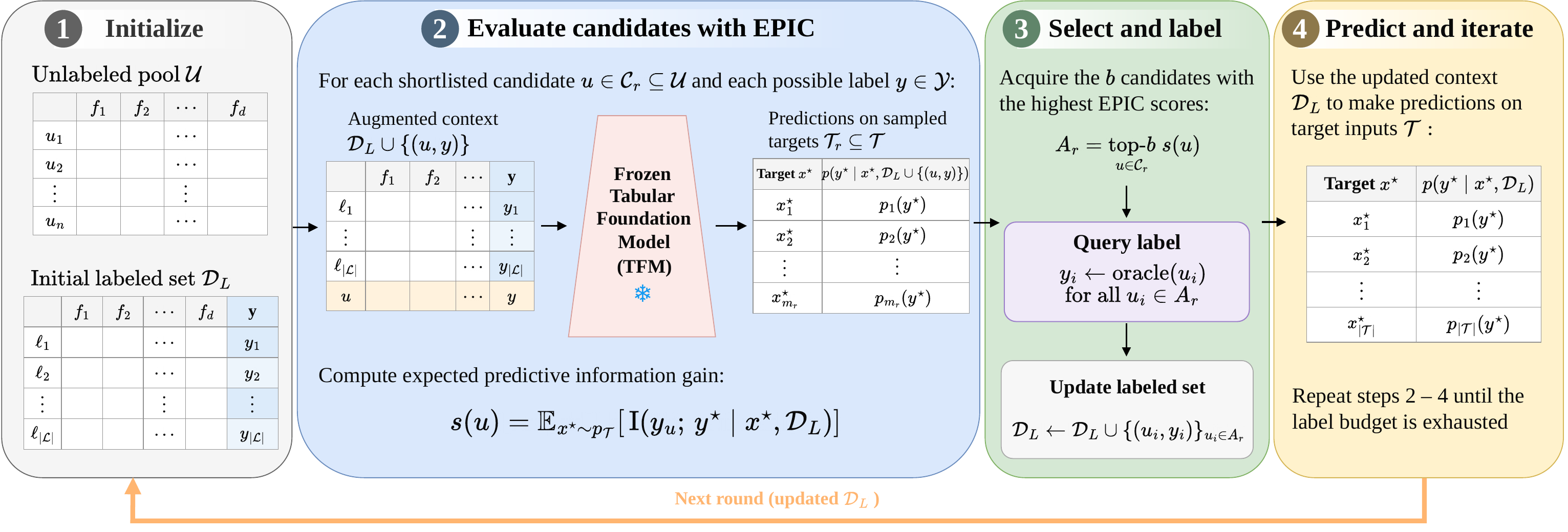}
    \caption{Workflow for the Expected Predictive Information Gain in context for tabular foundation models.}
    \label{fig:workflow}
\end{figure*}

We consider active in-context learning with a frozen TFM:
labels are bought from an oracle under a budget, 
and the models are affected only through their context.
To assess the acquisitions' impact, we compare three families: 
\emph{diversity-based} rules, which spread labels over the input space; \emph{uncertainty-based} rules, which label the candidates the model is least certain about; and \emph{target-aware} rules, which label candidates for the sake of the inputs on which the model will be used. 
We contribute two target-aware instruments: \textbf{TANN} (\cref{sec:method-targetnn}) labels, for the targets on which the model is least certain, their nearest pool rows; 
and \textbf{EPIC} (\cref{sec:method-epig}) scores a candidate by the information its label is expected to
carry about the target predictions. 

\subsection{Problem Setting}
\label{sec:method-setting}
Let $\Pool=\{x_i\}_{i=1}^{n}\subset\mathcal{X}$ be an unlabeled pool, and let an
oracle reveal the label $y_i\in\Y=\{1,\dots,C\}$ of a pool row at unit cost. Let
$\Tgt=\{x^\star_j\}_{j=1}^{m}$ be the target inputs whose labels are to be predicted,
distributed according to the deployment distribution $p_\Tgt$. We adopt the
transductive setting in which the target inputs, but not their labels, are
available before labeling starts. Otherwise, $\Tgt$ is a held-out sample of
unlabeled rows from $p_\Tgt$. For an index set $\Lab\subseteq[n]$ we write
$\Data_\Lab=\{(x_i,y_i)\}_{i\in\Lab}$.

A frozen TFM with parameters $\theta$ maps a labeled context $\Data$ and an input
$x$ to a predictive distribution $p_\theta(y\mid x,\Data)$ over $\Y$. 
Acquisition starts from an initial labeled set $\Lab_0$ that
contains one uniformly drawn row per class and proceeds in rounds: in round $r$ a batch $A_r\subseteq\Pool\setminus\Lab_r$ of $b$ rows is selected, their labels are queried, and we set $\Lab_{r+1} \leftarrow \Lab_r\cup A_r$, until the budget $|\Lab|=B$ is reached.
Predictions on $\Tgt$ are always made with the current context information defined in round $r$. 
The workflow is illustrated in \Cref{fig:workflow}.

\subsection{TANN: Target-aware nearest-neighbor acquisition}
\label{sec:method-targetnn}
One way to direct labels towards the targets is to place them next to targets on which the model is least certain. In round $r$, the model conditioned on $\Data_{\Lab_r}$ scores every target by its margin uncertainty
$u_r(x^\star)=1-\bigl(p_{(1)}-p_{(2)}\bigr)$, where $p_{(1)}\ge p_{(2)}$ are its two largest predicted class probabilities. Proximity is measured in the TFM's own row representation: for TabICL~\citep{qu2025tabicl}, $h(x;\Data)$ denotes the concatenated class tokens of row $x$ after the row-wise interaction stage, averaged over the ensemble members; with the normalization $\nu(v)=(v-\bar v\mathbf 1)/\lVert v-\bar v\mathbf 1\rVert_2$, the distance is $d_\Data(x,x')=\lVert\nu(h(x;\Data))-\nu(h(x';\Data))\rVert_2$. 
Because the column embedding of TabICL attends to the labeled context, this representation changes as labels are acquired. TANN visits the $b$ most uncertain targets in order of decreasing $u_r$ and assigns to each its nearest unlabeled pool row that has not been assigned yet,
\begin{align}
  \label{eq:targetnn}
  a_j=\argmin_{i\in\Pool_r\setminus 
  \{a_i\}_{i=1}^{i=j{-}1}} d_{\Data_{\Lab_r}}\bigl(x^\star_{\pi_r(j)},x_i\bigr),\quad j{=}1,\dots,b,
\end{align}
where $\pi_r$ orders the targets by decreasing uncertainty and
$\Pool_r=[n]\setminus\Lab_r$. A variant, dubbed TANN-feat,  uses the Euclidean distance of features standardized with pool statistics. 

\subsection{EPIC: \textbf{E}xpected \textbf{P}redictive \textbf{I}nformation Gain in \textbf{C}ontext}
\label{sec:method-epig}

The value of an observation for prediction is measured by the information it provides about the quantities to be predicted~\citep{lindley1956measure,
mackay1992information}. Prediction-oriented active learning 
leverages the expected predictive information gain (EPIG)~\citep{bickfordsmith2023epig}, i.e., the mutual information between the label of a candidate and the label of a target drawn from the target distribution. We instantiate it with the predictive model defined by a frozen TFM.

\begin{definition}[EPIC]
\label{def:icepig}
Given labeled data $\Data$, the \textit{\textbf{E}xpected \textbf{P}redictive \textbf{I}nformation Gain in \textbf{C}ontext}
of a candidate $x$ is
\begin{align}
  &\mathrm{EPIC}(x;\Data)
  \;=
  \;\E_{x^\star\sim p_\Tgt}\Bigl[
    \Ent\bigl[\bar p(y^\star\mid x^\star)\bigr]
    \;-\nonumber\\
  &\;\sum_{y\in\Y} p_\theta(y\mid x,\Data)\,
    \Ent\bigl[p_\theta(y^\star\mid x^\star,\Data\cup\{(x,y)\})\bigr]\Bigr], 
  \label{eq:icepig}
\end{align}
where $\bar p(y^\star\mid x^\star)=\sum_{y\in\Y}p_\theta(y\mid x,\Data)\, p_\theta(y^\star\mid x^\star,\Data\cup\{(x,y)\})$ is the predictive distribution of the target label marginalized over the unknown candidate label, and $\Ent$ denotes the Shannon entropy.
\end{definition}

The bracket in \cref{eq:icepig} is the mutual information between $y$ and $y^\star$ under the joint distribution
$p_\theta(y\mid x,\Data)\,p_\theta(y^\star\mid x^\star,\Data\cup\{(x,y)\})$. Therefore, it is non-negative and vanishes exactly when no possible label of $x$ changes the
prediction at $x^\star$. Intuitively, a candidate receives a high score when its possible labels would move the target predictions in different, confident directions, and a
low score when its label is already certain or irrelevant to the targets. Pseudo-code of EPIC is provided in \Cref{appdx:code}.\footnote{Code repository available on acceptance.}

\todo{discussion outcommented}

\subsection{Estimation with a frozen TFM}
\label{sec:method-estimation}

Evaluating \cref{eq:icepig} requires, for every candidate $x$ and every label $y\in\Y$, the predictions of the TFM on the targets with the augmented context $\Data\cup\{(x,y)\}$. With conventional models this would require refitting on every hypothetical dataset. Leveraging a TFM, it requires only a forward pass with one more context row. Each acquisition round estimates \cref{eq:icepig} as follows:

\textbf{Targets.}
The expectation over $p_\Tgt$ is replaced by its Monte Carlo estimate on a subsample
$\Tgt_r\subseteq\Tgt$ of size $m_r{=}\min(m,256)$, drawn uniformly without replacement and redrawn in every round. 

\textbf{Candidate shortlist.}
Every scored candidate requires $C$ hypothetical contexts, so we score a shortlist $\Cand_r$ of at most $s{=}64$ unlabeled rows (all unlabeled rows if fewer remain), composed of three parts of about equal size: the $\lfloor s/3\rfloor$ rows with the smallest margin between their two most probable classes; for the targets in $\Tgt_r$ with the smallest margins, their nearest unlabeled rows in the row-representation space of the TFM~\citep{qu2025tabicl}, until $\lfloor 2s/3\rfloor$ rows are collected; and uniformly drawn rows for the remainder, which keep the shortlist from being determined by the 
heuristics alone.

\textbf{Hypothetical contexts.}
For every $x\in\Cand_r$ and $y\in\Y$, the TFM is evaluated on $\Tgt_r$ with the context $\Data_{\Lab_r}\cup\{(x,y)\}$, which gives $s\cdot C$ predictive distributions for each target. 

\todo{outcommented}

With the true label of the candidate, a table reproduces exactly the prediction of the fitted estimator restricted to the context $\Lab_r\cup\{x\}$; for any label, it differs from an estimator fitted on $\Lab_r\cup\{x\}$ alone only through the preprocessing statistics, which are fitted on $\Lab_r\cup\Cand_r$. 
\todo{outcommented}

\textbf{Score.}
With $p_\theta(y\mid x,\Data_{\Lab_r})$ taken from the estimator fitted on $\Data_{\Lab_r}$, \cref{eq:icepig} is evaluated in closed form over $\Y$ and averaged over $\Tgt_r$. No sampling over labels is required.

\textbf{Cost.}
Let $c(k)$ be the cost of one forward pass on a table with $k$ rows and $M$ the number of ensemble members of the TFM ($M=8$ in our experiments), each of which processes every table. One round costs one fit and $O\bigl(s\,C\,M\,c(|\Lab_r|+1+m_r)\bigr)$ for the hypothetical contexts. Because $|\Lab_r|\le B$ and $m_r\le 256$, this cost is independent of the pool size, apart from the shortlist, which requires one pass over the unlabeled rows (ablation provided in \cref{appdx:epig-ablation}).

\textbf{Batch. } The batch $A_r$ consists of the $b$ shortlisted candidates with the largest scores. This greedy top-$b$ selection treats the candidates independently and may select near-duplicate rows whose information about the targets overlaps; joint batch criteria~\citep{kirsch2019batchbald} remove this redundancy at a higher cost. We use the independent selection throughout and leave joint batch criteria to future work.

\textbf{Scientific scope.}
We focus on classification, where the hypothetical labels can be enumerated. For regression, \cref{eq:icepig} can be estimated by drawing
hypothetical targets from $p_\theta(y\mid x,\Data)$ and replacing the entropies by those of the predictive distributions, which we leave to future work. 
All quantities are computed with forward passes of the frozen model, so the procedure applies to any TFM that permits context rows and their labels to be specified individually.

\section{Evaluation}
\label{sec:eval}

We organize the evaluation around the three findings: 
\textbf{When} diversity- and prediction-oriented acquisition are most valuable (\cref{sec:eval-when});
\textbf{Where} labels should be acquired (\cref{sec:eval-where}); and \textbf{How} EPIC performs as a
computation of prediction-oriented information gain (\cref{sec:eval-how}). \Cref{sec:eval-discussion}
discusses the limitations. Further results are reported in
\cref{sec:eval-secondary}.

\subsection{Evaluation protocol}
\label{sec:eval-protocol}

\textbf{Benchmarks.}
We use the twenty OpenML~\citep{vanschoren2014openml} classification datasets of Tab-AICL~\citep{treerath2026tabaicl} (details in \Cref{appdx:benchmarkDatasets}). They range from 145 to 581{,}012 rows, from 4 to 94 features and from 2 to 7 classes, and include strongly imbalanced tasks such as Blood-Transfusion, Bank-Marketing and JM1. Following Tab-AICL, datasets with more than 10{,}000 rows are subsampled uniformly to 10{,}000 rows. For every seed, each dataset is split into a pool of unlabeled candidates and a held-out test set by a stratified 70/30 split that is redrawn per seed. We refer to the eight datasets whose pools contain 1{,}358 to 7{,}000 rows as \emph{large} and to the remaining twelve (101 to 700 rows) as \emph{small}.

\textbf{Acquisition protocol.}
The context is initialized with one uniformly drawn row per class and grown in batches of $b{=}10$ rows until the budget $B{=}100$ is reached, or until half of the pool has been labeled on the smallest datasets. 
Every configuration is repeated with ten seeds, which change the split, the initial context and all random choices. 

\textbf{Competitors.}
We compare ten acquisition rules, which differ only in how the next batch is chosen, and group them into three families: \textbf{diversity-based}%
, 
\textbf{uncertainty-based} 
and
\textbf{target-aware}. 
Following \emph{Tab-AICL's rules}~\citep{treerath2026tabaicl}: Random (uniform sampling); Margin (smallest margin between the two most probable classes); Coreset (greedy $k$-center in the standardized feature space~\citep{sener2018coreset}); Hybrid (the most uncertain half of the pool by predictive entropy, then the row nearest to each of $b$ $k$-means centroids); and Proxy-Hybrid (a shortlist from a linear proxy model, then Hybrid on the shortlist). As \emph{Target-aware rules} (\cref{sec:method-targetnn}): TANN labels, for each of the $b$ target inputs with the smallest margin, its nearest unlabeled pool row in the TFM's row-representation space; TANN-feat measures the distance in the standardized feature space instead. 
Additionally, we implement Target-Random that assigns nearest pool rows to uniformly drawn targets, and Target-Coverage to the targets farthest from the labeled set; both use the target inputs but no uncertainty. Finally,  EPIC (\cref{sec:method-epig}) with $s{=}64$ shortlisted candidates and $m_r{=}256$ sampled targets per round. An ablation on $s$ and $m_r$ is provided in \cref{appdx:epig-ablation}.

\textbf{Baselines with other models.}
To relate the TabICL-based rules to the baselines of Tab-AICL, we further evaluate under the same protocol (\cref{tab:baselines}): the frozen TabPFN~v3 with the Tab-AICL rules (Random, Margin, Coreset, Hybrid and Proxy-Hybrid) and TANN-feat, the gradient-boosted tree models XGBoost-Margin and CatBoost-Margin, which are refitted on the labeled set in every round and select by their own margin, and
LabelSpreading-Random~\citep{treerath2026tabaicl}, a graph-based semi-supervised model
over the labeled rows, the pool and the test inputs with random acquisition.
In addition, to test whether the findings depend on the backbone, we repeat the complete study (all rules, including EPIC and TANN) with TabPFN~v3 as the frozen model (\cref{appdx:tabpfn}).

\textbf{Metrics.}
Following Tab-AICL, we measure the normalized area under the curve of Cohen's $\kappa$ on the test split over the acquisition rounds (AULC), which equals the average $\kappa$ over the trajectory and rewards rules that are accurate early. Because the purpose of acquisition is the model obtained once the budget is spent, we additionally report $\kappa$ at the budget (final $\kappa$) and use it to assess the models' capabilities at the budget. Cohen's $\kappa$ corrects accuracy for chance agreement, which matters on imbalanced datasets. 

\textbf{Statistical analysis.}
For every dataset, each metric is averaged over the ten seeds. Across the twenty datasets we report mean values and mean ranks, and for pairs of methods the number of wins, ties ($|\Delta|\le0.005$) and losses together with the two-sided Wilcoxon signed-rank test on the dataset means. When a method is compared with all nine others, we additionally apply the Holm correction for multiple comparisons and state it explicitly. All other $p$-values are uncorrected.

\textbf{Implementation.}
All acquisition rules of the main comparison use the same frozen TFM, TabICL with the publicly released v2 classification checkpoint~\citep{qu2025tabicl,qu2026tabiclv2}, with an ensemble of eight members and its default inference settings. 
The TabPFN baselines use the TabPFN~v3 classification checkpoints (binary and multiclass)~\citep{hollmann2025tabpfn}, also with eight ensemble members. XGBoost and CatBoost use their library defaults, as no validation data is available for tuning with at most 100 labels, and label spreading uses a seven-nearest-neighbor graph on standardized features. Features are passed to the model as provided by OpenML, with categorical columns ordinally encoded and missing values left to the model. 
Each job ran on one NVIDIA A40 GPU (48\,GB) with 16 CPU cores. 

\todo[]{there is a runtime analysis here}

\subsection{When: the value of acquisition criteria changes as the context grows}
\label{sec:eval-when}

\begin{figure}[h]
    \centering
    \includegraphics[width=\columnwidth]{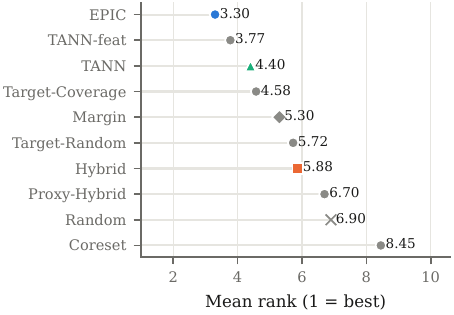}
    \caption{Mean rank of every acquisition rule over the twenty datasets (1 is best), for Cohen's $\kappa$ at the maximal label budget, with TabICL as the backbone.}
    \label{fig:ranks_final}
\end{figure}

\begin{figure}
    \centering
    \includegraphics[width=1\columnwidth]{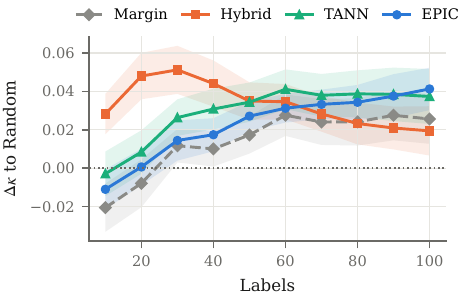}
    \caption{Learning curves on the fifteen datasets with a budget of 100 labels, with TabICL as the backbone. Difference to Random baselines where the bands show $\pm1$ standard error over
    datasets. }
    \label{fig:learning_curves_randomBaselines}
\end{figure}

\paragraph{Diversity leads in the cold start, prediction-oriented acquisition at the budget.}
\Cref{fig:ranks_final} illustrates that 
EPIC attains the highest $\kappa$ (0.578, rank 3.30) and improves on Hybrid (0.549) on 15 of the 20 datasets ($+0.030$, $p=0.011$; $p=0.064$ after Holm correction over nine comparisons), and TANN improves on Hybrid on 13 datasets ($+0.021$, $p=0.049$, uncorrected; \cref{tab:final,tab:pairwise}). 
The learning curves locate the transition (\cref{fig:learning_curves_randomBaselines}): Hybrid leads clearly during the first 40 labels, by up to $0.05$ $\kappa$ over Random, but its advantage shrinks steadily afterwards; TANN overtakes it at about 50 labels and EPIC at about 60, and both stay ahead until the budget. The figure averages the fifteen datasets whose budget is 100 labels, because averaging requires a common label axis and the five smallest datasets stop at half of their pool (50 to 74 labels); the AULC over all twenty datasets shows the same pattern. The learning curves after each acquisition round are provided in \Cref{appdx:when_learningCurves}.

\begin{table*}[t]
\centering
\small
\setlength{\tabcolsep}{3pt}
\caption{Cohen's $\kappa$ at the label budget (mean $\pm$ std over 10 seeds); best in bold, second underlined.}
\label{tab:final}
\resizebox{\textwidth}{!}{
\begin{tabular}{lcccccccccc}
\toprule
Dataset & \rotatebox{0}{Random} & \rotatebox{0}{Margin} & \rotatebox{0}{Coreset} & \rotatebox{0}{Hybrid} & \makecell{Proxy-\\Hybrid} & \makecell{Target-\\Random} & \makecell{Target-\\Coverage} & \rotatebox{0}{TANN-feat} & \rotatebox{0}{TANN} & \rotatebox{0}{EPIC} \\
\midrule
iris & \textbf{0.940\,{\scriptsize$\pm$0.053}} & 0.927\,{\scriptsize$\pm$0.051} & 0.920\,{\scriptsize$\pm$0.052} & 0.930\,{\scriptsize$\pm$0.041} & 0.930\,{\scriptsize$\pm$0.053} & 0.933\,{\scriptsize$\pm$0.047} & 0.933\,{\scriptsize$\pm$0.047} & \underline{0.940\,{\scriptsize$\pm$0.036}} & 0.930\,{\scriptsize$\pm$0.050} & 0.923\,{\scriptsize$\pm$0.052} \\
glass & 0.635\,{\scriptsize$\pm$0.058} & 0.651\,{\scriptsize$\pm$0.084} & 0.594\,{\scriptsize$\pm$0.070} & 0.623\,{\scriptsize$\pm$0.066} & 0.632\,{\scriptsize$\pm$0.090} & \underline{0.676\,{\scriptsize$\pm$0.046}} & 0.638\,{\scriptsize$\pm$0.065} & 0.675\,{\scriptsize$\pm$0.052} & 0.667\,{\scriptsize$\pm$0.080} & \textbf{0.706\,{\scriptsize$\pm$0.073}} \\
ionosphere & 0.878\,{\scriptsize$\pm$0.047} & 0.872\,{\scriptsize$\pm$0.028} & 0.740\,{\scriptsize$\pm$0.058} & 0.863\,{\scriptsize$\pm$0.039} & 0.864\,{\scriptsize$\pm$0.031} & 0.878\,{\scriptsize$\pm$0.048} & 0.833\,{\scriptsize$\pm$0.031} & 0.850\,{\scriptsize$\pm$0.051} & \underline{0.880\,{\scriptsize$\pm$0.030}} & \textbf{0.887\,{\scriptsize$\pm$0.040}} \\
balance-scale & 0.855\,{\scriptsize$\pm$0.039} & \textbf{0.975\,{\scriptsize$\pm$0.013}} & 0.899\,{\scriptsize$\pm$0.023} & 0.927\,{\scriptsize$\pm$0.015} & \underline{0.970\,{\scriptsize$\pm$0.015}} & 0.904\,{\scriptsize$\pm$0.031} & 0.922\,{\scriptsize$\pm$0.026} & 0.915\,{\scriptsize$\pm$0.036} & 0.897\,{\scriptsize$\pm$0.056} & 0.964\,{\scriptsize$\pm$0.019} \\
vehicle & 0.954\,{\scriptsize$\pm$0.028} & \underline{0.991\,{\scriptsize$\pm$0.009}} & 0.934\,{\scriptsize$\pm$0.034} & 0.965\,{\scriptsize$\pm$0.023} & 0.989\,{\scriptsize$\pm$0.011} & 0.957\,{\scriptsize$\pm$0.014} & 0.937\,{\scriptsize$\pm$0.041} & 0.986\,{\scriptsize$\pm$0.011} & 0.989\,{\scriptsize$\pm$0.006} & \textbf{0.996\,{\scriptsize$\pm$0.005}} \\
page-blocks & 0.695\,{\scriptsize$\pm$0.047} & 0.770\,{\scriptsize$\pm$0.074} & 0.695\,{\scriptsize$\pm$0.080} & 0.798\,{\scriptsize$\pm$0.025} & 0.791\,{\scriptsize$\pm$0.036} & 0.725\,{\scriptsize$\pm$0.030} & 0.804\,{\scriptsize$\pm$0.028} & \textbf{0.832\,{\scriptsize$\pm$0.023}} & \underline{0.819\,{\scriptsize$\pm$0.067}} & 0.812\,{\scriptsize$\pm$0.040} \\
parkinsons & 0.739\,{\scriptsize$\pm$0.119} & \underline{0.835\,{\scriptsize$\pm$0.100}} & 0.725\,{\scriptsize$\pm$0.115} & 0.799\,{\scriptsize$\pm$0.113} & 0.815\,{\scriptsize$\pm$0.107} & 0.763\,{\scriptsize$\pm$0.087} & 0.827\,{\scriptsize$\pm$0.096} & 0.833\,{\scriptsize$\pm$0.096} & 0.824\,{\scriptsize$\pm$0.120} & \textbf{0.844\,{\scriptsize$\pm$0.100}} \\
seeds & \underline{0.933\,{\scriptsize$\pm$0.032}} & 0.926\,{\scriptsize$\pm$0.038} & 0.929\,{\scriptsize$\pm$0.028} & 0.933\,{\scriptsize$\pm$0.030} & 0.926\,{\scriptsize$\pm$0.025} & 0.914\,{\scriptsize$\pm$0.032} & 0.929\,{\scriptsize$\pm$0.024} & \textbf{0.938\,{\scriptsize$\pm$0.036}} & 0.926\,{\scriptsize$\pm$0.033} & 0.929\,{\scriptsize$\pm$0.035} \\
bank-marketing & 0.235\,{\scriptsize$\pm$0.059} & 0.311\,{\scriptsize$\pm$0.063} & 0.279\,{\scriptsize$\pm$0.085} & 0.200\,{\scriptsize$\pm$0.103} & 0.292\,{\scriptsize$\pm$0.078} & \textbf{0.340\,{\scriptsize$\pm$0.076}} & 0.307\,{\scriptsize$\pm$0.088} & 0.254\,{\scriptsize$\pm$0.109} & \underline{0.321\,{\scriptsize$\pm$0.083}} & 0.252\,{\scriptsize$\pm$0.071} \\
adult & 0.441\,{\scriptsize$\pm$0.047} & 0.468\,{\scriptsize$\pm$0.037} & 0.442\,{\scriptsize$\pm$0.070} & 0.491\,{\scriptsize$\pm$0.035} & 0.464\,{\scriptsize$\pm$0.044} & 0.484\,{\scriptsize$\pm$0.045} & \underline{0.495\,{\scriptsize$\pm$0.032}} & 0.490\,{\scriptsize$\pm$0.027} & 0.487\,{\scriptsize$\pm$0.031} & \textbf{0.509\,{\scriptsize$\pm$0.038}} \\
covertype & 0.425\,{\scriptsize$\pm$0.027} & 0.446\,{\scriptsize$\pm$0.026} & 0.433\,{\scriptsize$\pm$0.049} & 0.436\,{\scriptsize$\pm$0.033} & 0.421\,{\scriptsize$\pm$0.050} & 0.435\,{\scriptsize$\pm$0.038} & 0.450\,{\scriptsize$\pm$0.036} & \textbf{0.459\,{\scriptsize$\pm$0.018}} & 0.445\,{\scriptsize$\pm$0.028} & \underline{0.458\,{\scriptsize$\pm$0.024}} \\
kc1 & 0.448\,{\scriptsize$\pm$0.155} & \textbf{0.511\,{\scriptsize$\pm$0.115}} & 0.392\,{\scriptsize$\pm$0.134} & 0.362\,{\scriptsize$\pm$0.148} & 0.431\,{\scriptsize$\pm$0.071} & 0.424\,{\scriptsize$\pm$0.133} & 0.465\,{\scriptsize$\pm$0.092} & 0.484\,{\scriptsize$\pm$0.105} & 0.454\,{\scriptsize$\pm$0.099} & \underline{0.508\,{\scriptsize$\pm$0.167}} \\
jm1 & \textbf{0.173\,{\scriptsize$\pm$0.053}} & 0.090\,{\scriptsize$\pm$0.062} & 0.116\,{\scriptsize$\pm$0.049} & \underline{0.162\,{\scriptsize$\pm$0.042}} & 0.101\,{\scriptsize$\pm$0.060} & 0.128\,{\scriptsize$\pm$0.067} & 0.136\,{\scriptsize$\pm$0.065} & 0.055\,{\scriptsize$\pm$0.065} & 0.071\,{\scriptsize$\pm$0.067} & 0.123\,{\scriptsize$\pm$0.054} \\
blood-transfusion & 0.152\,{\scriptsize$\pm$0.088} & 0.213\,{\scriptsize$\pm$0.071} & 0.211\,{\scriptsize$\pm$0.076} & \underline{0.255\,{\scriptsize$\pm$0.068}} & 0.176\,{\scriptsize$\pm$0.113} & 0.246\,{\scriptsize$\pm$0.097} & 0.228\,{\scriptsize$\pm$0.096} & 0.239\,{\scriptsize$\pm$0.112} & \textbf{0.266\,{\scriptsize$\pm$0.082}} & 0.202\,{\scriptsize$\pm$0.121} \\
diabetes & 0.412\,{\scriptsize$\pm$0.075} & 0.406\,{\scriptsize$\pm$0.060} & 0.422\,{\scriptsize$\pm$0.049} & 0.421\,{\scriptsize$\pm$0.056} & 0.419\,{\scriptsize$\pm$0.051} & 0.425\,{\scriptsize$\pm$0.037} & 0.433\,{\scriptsize$\pm$0.049} & \underline{0.446\,{\scriptsize$\pm$0.052}} & \textbf{0.448\,{\scriptsize$\pm$0.045}} & 0.438\,{\scriptsize$\pm$0.049} \\
tic-tac-toe & 0.288\,{\scriptsize$\pm$0.036} & 0.295\,{\scriptsize$\pm$0.059} & 0.215\,{\scriptsize$\pm$0.065} & 0.209\,{\scriptsize$\pm$0.051} & 0.259\,{\scriptsize$\pm$0.050} & 0.270\,{\scriptsize$\pm$0.059} & \textbf{0.308\,{\scriptsize$\pm$0.040}} & \underline{0.301\,{\scriptsize$\pm$0.068}} & 0.291\,{\scriptsize$\pm$0.059} & 0.291\,{\scriptsize$\pm$0.049} \\
credit-g & 0.259\,{\scriptsize$\pm$0.100} & 0.265\,{\scriptsize$\pm$0.059} & 0.251\,{\scriptsize$\pm$0.069} & \textbf{0.309\,{\scriptsize$\pm$0.063}} & 0.279\,{\scriptsize$\pm$0.056} & 0.247\,{\scriptsize$\pm$0.071} & 0.248\,{\scriptsize$\pm$0.090} & 0.280\,{\scriptsize$\pm$0.085} & 0.271\,{\scriptsize$\pm$0.055} & \underline{0.289\,{\scriptsize$\pm$0.074}} \\
steel-plates & 0.560\,{\scriptsize$\pm$0.022} & 0.577\,{\scriptsize$\pm$0.030} & 0.546\,{\scriptsize$\pm$0.061} & 0.585\,{\scriptsize$\pm$0.023} & 0.537\,{\scriptsize$\pm$0.072} & 0.577\,{\scriptsize$\pm$0.042} & 0.595\,{\scriptsize$\pm$0.050} & 0.619\,{\scriptsize$\pm$0.022} & \underline{0.626\,{\scriptsize$\pm$0.033}} & \textbf{0.627\,{\scriptsize$\pm$0.037}} \\
phoneme & 0.545\,{\scriptsize$\pm$0.031} & 0.609\,{\scriptsize$\pm$0.031} & 0.523\,{\scriptsize$\pm$0.028} & 0.540\,{\scriptsize$\pm$0.055} & 0.524\,{\scriptsize$\pm$0.155} & 0.563\,{\scriptsize$\pm$0.035} & 0.585\,{\scriptsize$\pm$0.031} & 0.614\,{\scriptsize$\pm$0.038} & \textbf{0.626\,{\scriptsize$\pm$0.028}} & \underline{0.623\,{\scriptsize$\pm$0.031}} \\
ilpd & 0.163\,{\scriptsize$\pm$0.062} & 0.128\,{\scriptsize$\pm$0.062} & 0.143\,{\scriptsize$\pm$0.095} & 0.162\,{\scriptsize$\pm$0.054} & 0.183\,{\scriptsize$\pm$0.087} & \textbf{0.230\,{\scriptsize$\pm$0.076}} & \underline{0.216\,{\scriptsize$\pm$0.105}} & 0.165\,{\scriptsize$\pm$0.057} & 0.156\,{\scriptsize$\pm$0.055} & 0.181\,{\scriptsize$\pm$0.077} \\
\midrule
Mean rank $\downarrow$ & 6.90 & 5.30 & 8.45 & 5.88 & 6.70 & 5.72 & 4.58 & \underline{3.77} & 4.40 & \textbf{3.30} \\
\bottomrule
\end{tabular}
}
\end{table*}

\textbf{The transition recurs with a second model and at larger budgets.}
The full replication with TabPFN as the backbone (\cref{appdx:tabpfn}) shows the same transition: Hybrid leads over the trajectory (AULC 0.456 vs.\ 0.447 for EPIC and 0.450 for TANN; ahead on 13 and 10 of 20 datasets, $p=0.33$ and $p=0.43$), whereas EPIC and TANN lead at the budget (final $\kappa$ 0.558 and 0.554 vs.\ 0.534; 14/1/5, $p=0.017$, and 12/4/4, $p=0.040$), with EPIC overtaking Hybrid after about 70 labels (\cref{fig:tabpfn-curves}).
With TabICL, when the budget is expressed as a fraction of the pool on the eight datasets with large pools (\cref{tab:reduction}), Hybrid retains the most full-context $\kappa$ with 1\% of the pool labeled (60.4\% vs.\ 50.9\% for EPIC, ahead on 7 of 8 datasets, $p=0.023$), the two are level at 2\%, and EPIC leads at 5\% and 10\% (80.3\% vs.\ 76.0\% and 87.0\% vs.\ 82.0\%, 7/0/1 datasets, $p=0.078$ and $p=0.055$).

With very few labels, the frozen model's predictions on the targets carry little information, so criteria that rely on them, its uncertainty and its response to hypothetical labels, are unreliable, whereas spreading the first labels over the input space is effective regardless. As the context grows, the predictions become informative enough to direct acquisition, and diversity loses its value. This mirrors the budget dependence that~\citet{hacohen2022budget} report for retrained models, here for a learner that is \textit{never trained} and \textit{changes only} through its context. 

\subsection{Where: target awareness matters substantially}
\label{sec:eval-where}

\textbf{Using the target inputs helps, and so does choosing which targets to serve.}
The target-aware rules see the test inputs, which the Tab-AICL rules do not. The two controls separate the use of these inputs from the choice of the targets: they assign nearest pool rows to targets drawn at random (Target-Random) or chosen for coverage (Target-Coverage), without uncertainty. Access to the target inputs alone already improves on random labeling: Target-Random in AULC on 13 of 20 datasets ($+0.016$, $p=0.008$) and Target-Coverage at the budget on 13 ($+0.028$, $p=0.012$). 
Choosing the targets by the model's uncertainty or by information gain improves further: at the budget, TANN improves on Target-Random on 12 datasets ($+0.014$, $p=0.070$), and EPIC on 15 ($+0.022$, $p=0.036$) and on Target-Coverage on 13 ($+0.014$, $p=0.091$)~ 
(details in \Cref{tab:aulc,tab:final}).

\textbf{Target awareness captures most of the benefit.}
TANN and EPIC are statistically indistinguishable, in AULC (8/5/7 datasets, $p=0.96$) and at the budget (11/4/5 in favor of EPIC, $+0.008$, $p=0.13$), although TANN needs no model of how a label changes the predictions and is roughly eight times cheaper (18\,s vs.\ 145\,s per run on the large datasets). 
Nor does the geometry in which proximity is measured matter: TANN in the TFM's row representation and in the standardized feature space are indistinguishable (AULC 6/6/8, $p=0.84$; final $\kappa$ 6/3/11, $p=0.60$). 
TANN also saves labels relative to random labeling about as much as EPIC (\cref{sec:eval-how}: median 61\% vs.\ 63\% of the budget for the quality Random reaches with all of it, compared with 78\% for Margin and Hybrid), and its labels transfer to tree-based models almost as well (\cref{tab:transfer}). 
Therefore, serving uncertain targets with nearby labels captures most of what prediction-oriented acquisition achieves.
With TabPFN, TANN and EPIC again improve on both target-aware controls, but target access alone does not: Target-Random is level with Random ($p=0.39$; \cref{appdx:tabpfn}).

\subsection{How: EPIC computes prediction-oriented information gain in context}
\label{sec:eval-how}

\textbf{Best models at the budget.}
EPIC attains the highest mean $\kappa$ at the budget and the best mean rank (\cref{tab:final}). In pairwise comparisons (\cref{tab:pairwise}), it improves on Random on 16 of 20 datasets ($+0.042$), on Coreset on 16 ($+0.058$) and on Proxy-Hybrid on 15 ($+0.028$), all significant after Holm correction, and on Margin, the best Tab-AICL rule at the budget, on 13 ($+0.015$, $p=0.024$ uncorrected).
With TabPFN as the backbone, EPIC again attains the best final $\kappa$ (14/1/5 against Hybrid, $p=0.017$, Holm-adjusted $p=0.086$), and the rankings of the rules agree across backbones (Spearman $\rho=0.89$; \cref{tab:tabpfn-vs-tabicl}).
A comparison against baselines of other competitors (TabPFN~v3, XGBoost, CatBoost) is provided in \cref{appdx:how_baselines}.
Reshaping the context of the acquired labels improves $\kappa$ by at most 0.015 (confident pseudo-labels) and restricting it to local contexts lowers $\kappa$, whereas the ordering of the rules remains unchanged (\cref{appdx:context}).

\begin{table}[t]
\centering
\small
\setlength{\tabcolsep}{3pt}
\caption{Pairwise comparison across the 20 datasets: wins/ties/losses of the row method against each rule (tie: $|\Delta|\le0.005$), mean difference $\Delta$ of the dataset means, and the $p$-value of a two-sided Wilcoxon signed-rank test ($p<0.05$ in bold).}
\label{tab:pairwise}
\resizebox{\columnwidth}{!}{
\begin{tabular}{llcccccc}
\toprule
& & \multicolumn{3}{c}{AULC $\kappa$} & \multicolumn{3}{c}{Final $\kappa$} \\
\cmidrule(lr){3-5}\cmidrule(lr){6-8}
Method & vs. & W/T/L & $\Delta$ & $p$ & W/T/L & $\Delta$ & $p$ \\
\midrule
EPIC & Random & 16/1/3 & +0.026 & \textbf{0.001} & 16/2/2 & +0.042 & \textbf{0.000} \\
 & Margin & 13/4/3 & +0.012 & \textbf{0.017} & 13/4/3 & +0.015 & \textbf{0.024} \\
 & Coreset & 16/0/4 & +0.055 & \textbf{0.000} & 16/2/2 & +0.058 & \textbf{0.001} \\
 & Hybrid & 7/2/11 & -0.003 & 0.701 & 15/1/4 & +0.030 & \textbf{0.011} \\
 & Proxy-Hybrid & 10/4/6 & +0.014 & 0.177 & 15/2/3 & +0.028 & \textbf{0.001} \\
 & Target-Random & 10/6/4 & +0.010 & 0.165 & 15/0/5 & +0.022 & \textbf{0.036} \\
 & Target-Coverage & 11/4/5 & +0.008 & 0.277 & 13/1/6 & +0.014 & 0.091 \\
 & TANN-feat & 9/4/7 & +0.003 & 0.927 & 12/2/6 & +0.009 & 0.165 \\
 & TANN & 8/5/7 & +0.000 & 0.956 & 11/4/5 & +0.008 & 0.133 \\
\midrule
TANN & Random & 16/2/2 & +0.026 & \textbf{0.002} & 14/2/4 & +0.033 & \textbf{0.006} \\
 & Margin & 15/1/4 & +0.012 & \textbf{0.015} & 11/5/4 & +0.006 & 0.198 \\
 & Coreset & 18/1/1 & +0.055 & \textbf{0.000} & 17/2/1 & +0.049 & \textbf{0.000} \\
 & Hybrid & 7/3/10 & -0.004 & 0.571 & 13/2/5 & +0.021 & \textbf{0.049} \\
 & Proxy-Hybrid & 9/5/6 & +0.013 & 0.154 & 13/3/4 & +0.020 & \textbf{0.035} \\
 & Target-Random & 12/3/5 & +0.010 & 0.083 & 12/3/5 & +0.014 & 0.070 \\
 & Target-Coverage & 10/4/6 & +0.007 & 0.261 & 10/4/6 & +0.005 & 0.409 \\
 & TANN-feat & 6/6/8 & +0.003 & 0.841 & 6/3/11 & +0.001 & 0.596 \\
 & EPIC & 7/5/8 & -0.000 & 0.956 & 5/4/11 & -0.008 & 0.133 \\
\bottomrule
\end{tabular}
}
\end{table}

\begin{figure}[t]
  \centering
  \includegraphics[width=\linewidth]{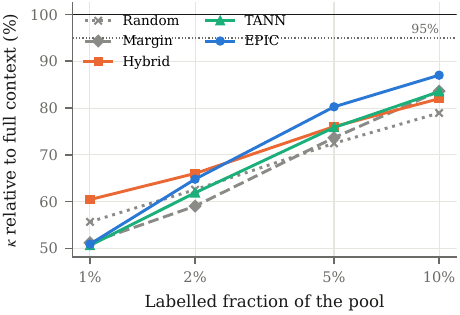}
  \caption{Cohen's $\kappa$ with a fraction of the pool labeled, TabICL as the backbone, relative to the fully labeled pool, on the eight datasets with pools of 1{,}358 to 7{,}000 rows
    (mean over datasets of the means over ten seeds). The dotted line marks 95\% of the full-context performance.}
  \label{fig:reduction}
\end{figure}

\textbf{Fewer labels for the same quality.}
Relative to random labeling, EPIC reaches the $\kappa$ that Random attains with the full budget using a median of 63\% of the labels (mean 64\%), and no more than 80\% on 14 of the 20 datasets, determined from the seed-averaged learning curves. Relative to the fully labeled pool, we extended the budget to 10\% of the pool in batches of 0.5\% on the eight datasets whose pools contain 1{,}358 to 7{,}000 rows, where a fraction of a few percent is still meaningful, and compared $\kappa$ at 1\%, 2\%, 5\% and 10\% with the full-context $\kappa$ on the same splits over ten seeds (\cref{fig:reduction}). 
With 5\% and 10\% of the pool labeled, EPIC already retains on average 80.3\% and 87.0\% of the full-context $\kappa$, compared with 72.5\% and 78.9\% for Random, and improves on Random on all eight datasets at both fractions ($p=0.008$). 
More details are provided in \Cref{appdx:how_reduction}.
EPIC also provides a signal for when to stop labeling: the largest information gain of a round decreases steadily as labels are acquired, from about $10^{-1}$ to about $2\cdot10^{-3}$ nats over the budget. Stopping a run once it falls below a threshold $\tau$, however, improves only marginally on stopping every run at the same fraction of its budget (by at most 0.010 $\kappa$; \cref{appdx:cost}).


\textbf{Labels that transfer to other models.}
A rule that exploits the frozen TFM might select labels that are useful only to that model. We therefore re-ran five TabICL rules and, after every round, also fitted XGBoost and CatBoost on the labels acquired so far (\cref{tab:transfer}). 
Trained on the labels selected by EPIC, XGBoost and CatBoost reach a final $\kappa$ of 0.530 and 0.543, compared with 0.498 and 0.525 when they acquire their own labels by margin (17/1/2 datasets, $p<0.001$, and 12/2/6, $p=0.030$), and improve on randomly acquired labels on 15 datasets for both models ($p\le0.003$). XGBoost reaches the $\kappa$ of its own acquisition at the budget with a median of 73\% of the labels selected by EPIC, on 19 of the 20 datasets; CatBoost, whose own acquisition is stronger, needs 75\% on average and reaches it on 12. 

\begin{table}[t]
\centering
\small
\caption{Label transfer: final $\kappa$ of models trained on the labels acquired by each TabICL rule (mean over datasets and seeds). The diagonal reference rows give the same model with its own margin-based acquisition. The rules were rerun for this analysis; the $\kappa$ of EPIC differs from \cref{tab:final} (0.578) because of GPU nondeterminism.}
\label{tab:transfer}
\resizebox{\columnwidth}{!}{
\begin{tabular}{lccc}
\toprule
Labels acquired by & TabICL & CatBoost & XGBoost \\
\midrule
Random & 0.536 & 0.500 & 0.487 \\
Margin & 0.563 & 0.523 & 0.514 \\
Hybrid & 0.549 & 0.519 & 0.504 \\
TANN & \underline{0.570} & \underline{0.538} & \underline{0.519} \\
EPIC & \textbf{0.579} & \textbf{0.543} & \textbf{0.530} \\
\midrule
CatBoost-Margin (own labels) & -- & 0.525 & -- \\
XGBoost-Margin (own labels) & -- & -- & 0.498 \\
\bottomrule
\end{tabular}
}
\end{table}

\subsection{Discussion and limitations}
\label{sec:eval-discussion}

Acquisition for a frozen TFM is not a single fixed problem. Early on, the value of a label
lies in covering the input space; once the model's predictions are informative, it lies in
resolving the predictions where the model will be used, and serving uncertain targets with
nearby labels captures most of this value. EPIC expresses the latter principle exactly and yields the best models at the budget, while TANN offers most of its benefit at a
fraction of its cost.

\textbf{Limitations.}
(i) \emph{No adaptive rule yet.} We observe when each criterion is valuable but do not
evaluate a rule that switches between them; whether such a switch improves on both, and
how to detect the transition, is the most direct test of the first finding. 
(ii) \emph{Transductive setting.} The target-aware rules require the target inputs, or a representative sample of them, before labeling. 
(iii) \emph{Classification.} EPIC enumerates hypothetical labels and is implemented for
classification only. 
(iv) \emph{Approximations.} EPIC selects batches greedily, so the rows of a batch may carry overlapping information. 
Its two other approximations, the shortlist of candidates and the sample of targets, do not affect $\kappa$ at the budget (\cref{appdx:epig-ablation}) in our setting. 


\section{Conclusion}
\label{sec:conclusion}

We ask how the value of diversity, predictive uncertainty and target-aware information gain changes as a frozen tabular foundation model (TFM) acquires more context. Across twenty benchmarks and two TFMs, the answer has three parts. Diversity is the best use of the first few dozen labels, and prediction-oriented acquisition overtakes it once the model's predictions become informative. Where the model will be used matters substantially: serving the targets on which the model is least certain with nearby labels captures most of the benefit of prediction-oriented acquisition, independently of how proximity is measured. And in-context conditioning makes prediction-oriented information gain computable for a frozen model: EPIC yields the best models at the label budget, 
retains on average 80--87\% of the chance-corrected full-data performance with 5--10\% of the pool labeled on the eight benchmarks with the largest pools, and selects labels that also improve gradient-boosted trees.

\section*{AI use statement}
In this work, we used generative AI tools to assist with code modifications, generate scripts for higher-quality figure plotting. We have not used generative AI tools for research ideas, designing core methodology, or writing initial text from scratch. Additionally, we used generative AI tools to improve sentence fluency while still using our original wording and arguments. We have reviewed all AI-assisted work: all modified code and plotting scripts were manually inspected, executed, and verified for correctness by the authors, and all edited text was carefully checked to ensure it had the same semantic meaning as our original version. We take responsibility for the final content of this work, including text, claims or artifacts produced with the aid of generative AI.

\section*{Acknowledgments}
We acknowledge funding from the \emph{Bayerisches Staatsministerium für Wirtschaft, Landesentwicklung und Energie} under the research programme \emph{BayVFP Förderlinie Digitalisierung} (grant reference DIK-2507-0024// DIK0836/01).

We gratefully acknowledge the scientific support and HPC resources provided by the Erlangen National High Performance Computing Center (NHR@FAU) of the Friedrich-Alexander-Universität Erlangen-Nürnberg (FAU) under the NHR project DeepSmall - Meta-learning for regularizing deep networks under small data regimes. NHR funding is provided by federal and Bavarian state authorities. NHR@FAU hardware is partially funded by the German Research Foundation (DFG) – 440719683.

\section*{Impact Statement}
This paper presents work whose goal is to advance the field
of Machine Learning. There are many potential societal
consequences of our work, none which we feel must be
specifically highlighted here.

\bibliographystyle{icml2026}
\bibliography{references}

@inproceedings{hollmann2023tabpfn,
  title={TabPFN: A Transformer That Solves Small Tabular Classification Problems in a Second}, 
  author={Noah Hollmann and Samuel Müller and Katharina Eggensperger and Frank Hutter},
  booktitle = {International Conference on Learning Representations (ICLR)},
  year      = {2023},
  note      = {arXiv:2207.01848},
}

@article{hollmann2025tabpfn,
  title = {Accurate predictions on small data with a tabular foundation model},
  volume = {637},
  ISSN = {1476-4687},
  url = {http://dx.doi.org/10.1038/s41586-024-08328-6},
  DOI = {10.1038/s41586-024-08328-6},
  number = {8045},
  journal = {Nature},
  publisher = {Springer Science and Business Media LLC},
  author = {Hollmann,  Noah and M\"{u}ller,  Samuel and Purucker,  Lennart and Krishnakumar,  Arjun and K\"{o}rfer,  Max and Hoo,  Shi Bin and Schirrmeister,  Robin Tibor and Hutter,  Frank},
  year = {2025},
  month = Jan,
  pages = {319–326}
}

@inproceedings{qu2025tabicl,
   title={TabICL: A Tabular Foundation Model for In-Context Learning on Large Data}, 
  author={Jingang Qu and David Holzmüller and Gaël Varoquaux and Marine Le Morvan},
  booktitle = {Proceedings of the 42nd International Conference on Machine Learning (ICML)},
  year      = {2025},
  note      = {arXiv:2502.05564},
}

@inproceedings{qu2026tabiclv2,
  title={TabICLv2: A better, faster, scalable, and open tabular foundation model}, 
  author={Jingang Qu and David Holzmüller and Gaël Varoquaux and Marine Le Morvan},
  booktitle = {Proceedings of the 43rd International Conference on Machine Learning (ICML)},
  year      = {2026},
  note      = {arXiv:2602.11139},
}

@article{treerath2026tabaicl,
      title={Active In-Context Learning for Tabular Foundation Models}, 
      author={Wilailuck Treerath and Fabrizio Pittorino},
      year={2026},
      eprint={2603.27385},
      archivePrefix={arXiv},
      primaryClass={cs.LG},
      url={https://arxiv.org/abs/2603.27385}, 
}

@article{chen2026vipcop,
      title={VIP-COP: Context Optimization for Tabular Foundation Models}, 
      author={Yilong Chen and Xueying Ding and Leman Akoglu},
      year={2026},
      eprint={2605.12904},
      archivePrefix={arXiv},
      primaryClass={cs.LG},
      url={https://arxiv.org/abs/2605.12904}, 
}

@article{ma2024icd,
      title={In-Context Data Distillation with TabPFN}, 
      author={Junwei Ma and Valentin Thomas and Guangwei Yu and Anthony Caterini},
      year={2024},
      eprint={2402.06971},
      archivePrefix={arXiv},
      primaryClass={cs.LG},
      url={https://arxiv.org/abs/2402.06971}, 
}

@inproceedings{feuer2024tunetables,
  author = {Feuer, Benjamin and Schirrmeister, Robin Tibor and Cherepanova, Valeriia and Hegde, Chinmay and Hutter, Frank and Goldblum, Micah and Cohen, Niv and White, Colin}, 
  title = {TuneTables: context optimization for scalable prior-data fitted networks}, 
  year = {2024}, 
  isbn = {9798331314385}, 
  publisher = {Curran Associates Inc.}, 
  address = {Red Hook, NY, USA}, 
  booktitle = {Proceedings of the 38th International Conference on Neural Information Processing Systems}, articleno = {2654}, 
  numpages = {35}, 
  location = {Vancouver, BC, Canada}, series = {NIPS '24}
}

@article{thomas2024localpfn,
author = {Thomas, Valentin and Ma, Junwei and Hosseinzadeh, Rasa and Golestan, Keyvan and Yu, Guangwei and Volkovs, Maksims and Caterini, Anthony}, 
title = {Retrieval \& fine-tuning for in-context tabular models}, year = {2024}, 
isbn = {9798331314385}, 
publisher = {Curran Associates Inc.}, 
address = {Red Hook, NY, USA}, 
booktitle = {Proceedings of the 38th International Conference on Neural Information Processing Systems}, articleno = {3442}, 
numpages = {29}, location = {Vancouver, BC, Canada}, series = {NIPS '24}
}

@article{guta2026localized,
     title={Localized TabICLv2: Scaling Tabular In-Context Learning through k-NN}, 
      author={Beimnet Bekele Guta},
      year={2026},
      eprint={2608.16429},
      archivePrefix={arXiv},
      primaryClass={cs.LG},
      url={https://arxiv.org/abs/2608.16429}, 
  note    = {2nd ICML Workshop on Foundation Models for Structured Data (FMSD)},
}

@techreport{settles2009active,
  author = {Settles, Burr},
  institution = {University of Wisconsin--Madison},
  number = 1648,
  title = {Active Learning Literature Survey},
  type = {Computer Sciences Technical Report},
  url = {http://axon.cs.byu.edu/~martinez/classes/778/Papers/settles.activelearning.pdf},
  year = 2009
}

@article{houlsby2011bald,
  author = {Houlsby, Neil and Huszar, Ferenc and Ghahramani, Zoubin and Lengyel, Máté},
  ee = {https://doi.org/10.48550/arXiv.1112.5745},
  journal = {CoRR},
  title = {Bayesian Active Learning for Classification and Preference Learning},
  url = {http://dblp.uni-trier.de/db/journals/corr/corr1112.html#abs-1112-5745},
  volume = {abs/1112.5745},
  year = 2011
}

@inproceedings{sener2018coreset,
  author       = {Ozan Sener and
                  Silvio Savarese},
  title        = {Active Learning for Convolutional Neural Networks: {A} Core-Set Approach},
  booktitle    = {6th International Conference on Learning Representations, {ICLR} 2018,
                  Vancouver, BC, Canada, April 30 - May 3, 2018, Conference Track Proceedings},
  publisher    = {OpenReview.net},
  year         = {2018},
  url          = {https://openreview.net/forum?id=H1aIuk-RW},
  bibsource    = {dblp computer science bibliography, https://dblp.org}
}

@inproceedings{yu2006transductive,
author = {Yu, Kai and Bi, Jinbo and Tresp, Volker}, title = {Active learning via transductive experimental design}, year = {2006}, isbn = {1595933832}, publisher = {Association for Computing Machinery}, address = {New York, NY, USA}, url = {https://doi.org/10.1145/1143844.1143980}, doi = {10.1145/1143844.1143980}, booktitle = {Proceedings of the 23rd International Conference on Machine Learning}, pages = {1081–1088}, numpages = {8}, location = {Pittsburgh, Pennsylvania, USA}, series = {ICML '06}
}

@article{rubachev2024tabred,
      title={TabReD: Analyzing Pitfalls and Filling the Gaps in Tabular Deep Learning Benchmarks}, 
      author={Ivan Rubachev and Nikolay Kartashev and Yury Gorishniy and Artem Babenko},
      year={2024},
      eprint={2406.19380},
      archivePrefix={arXiv},
      primaryClass={cs.LG},
      url={https://arxiv.org/abs/2406.19380}, 
}

@article{vanschoren2014openml,
author = {Vanschoren, Joaquin and van Rijn, Jan N. and Bischl, Bernd and Torgo, Luis}, title = {OpenML: networked science in machine learning}, year = {2014}, issue_date = {December 2013}, publisher = {Association for Computing Machinery}, address = {New York, NY, USA}, volume = {15}, number = {2}, issn = {1931-0145}, url = {https://doi.org/10.1145/2641190.2641198}, doi = {10.1145/2641190.2641198}, journal = {SIGKDD Explor. Newsl.}, month = jun, pages = {49–60}, numpages = {12}
}

@inproceedings{bickfordsmith2023epig,
  author       = {Freddie Bickford Smith and
                  Andreas Kirsch and
                  Sebastian Farquhar and
                  Yarin Gal and
                  Adam Foster and
                  Tom Rainforth},
  editor       = {Francisco J. R. Ruiz and
                  Jennifer G. Dy and
                  Jan{-}Willem van de Meent},
  title        = {Prediction-Oriented Bayesian Active Learning},
  booktitle    = {International Conference on Artificial Intelligence and Statistics,
                  25-27 April 2023, Palau de Congressos, Valencia, Spain},
  series       = {Proceedings of Machine Learning Research},
  volume       = {206},
  pages        = {7331--7348},
  publisher    = {{PMLR}},
  year         = {2023},
  url          = {https://proceedings.mlr.press/v206/smith23a.html},
  bibsource    = {dblp computer science bibliography, https://dblp.org}
}

@inproceedings{kirsch2019batchbald,
title = {BatchBALD: efficient and diverse batch acquisition for deep Bayesian active learning}, author={Andreas Kirsch and Joost van Amersfoort and Yarin Gal}, year = {2019}, publisher = {Curran Associates Inc.}, address = {Red Hook, NY, USA}, booktitle = {Proceedings of the 33rd International Conference on Neural Information Processing Systems}, articleno = {631}, numpages = {12}
}

@article{mackay1992information,
author = {MacKay, David J. C.}, title = {Information-based objective functions for active data selection}, year = {1992}, issue_date = {July 1992}, publisher = {MIT Press}, address = {Cambridge, MA, USA}, volume = {4}, number = {4}, issn = {0899-7667}, url = {https://doi.org/10.1162/neco.1992.4.4.590}, doi = {10.1162/neco.1992.4.4.590}, journal = {Neural Comput.}, month = jul, pages = {590–604}, numpages = {15}
}

@article{lindley1956measure,
  title={On a Measure of the Information Provided by an Experiment},
  author={Dennis V. Lindley},
  journal={Annals of Mathematical Statistics},
  year={1956},
  volume={27},
  pages={986-1005},
  url={https://api.semanticscholar.org/CorpusID:123582195}
}

@inproceedings{hacohen2022budget,
  title = 	 {Active Learning on a Budget: Opposite Strategies Suit High and Low Budgets},
  author =       {Hacohen, Guy and Dekel, Avihu and Weinshall, Daphna},
  booktitle = 	 {Proceedings of the 39th International Conference on Machine Learning},
  pages = 	 {8175--8195},
  year = 	 {2022},
  editor = 	 {Chaudhuri, Kamalika and Jegelka, Stefanie and Song, Le and Szepesvari, Csaba and Niu, Gang and Sabato, Sivan},
  volume = 	 {162},
  series = 	 {Proceedings of Machine Learning Research},
  month = 	 {17--23 Jul},
  publisher =    {PMLR},
  url = 	 {https://proceedings.mlr.press/v162/hacohen22a.html},
}

@inproceedings{dong_2024_iclsurvey,
    title = "A Survey on In-context Learning",
    author = "Dong, Qingxiu  and
      Li, Lei  and
      Dai, Damai  and
      Zheng, Ce  and
      Ma, Jingyuan  and
      Li, Rui  and
      Xia, Heming  and
      Xu, Jingjing  and
      Wu, Zhiyong  and
      Chang, Baobao  and
      Sun, Xu  and
      Li, Lei  and
      Sui, Zhifang",
    editor = "Al-Onaizan, Yaser  and
      Bansal, Mohit  and
      Chen, Yun-Nung",
    booktitle = "Proceedings of the 2024 Conference on Empirical Methods in Natural Language Processing",
    month = nov,
    year = "2024",
    address = "Miami, Florida, USA",
    publisher = "Association for Computational Linguistics",
    url = "https://aclanthology.org/2024.emnlp-main.64/",
    doi = "10.18653/v1/2024.emnlp-main.64",
    pages = "1107--1128"
}

@article{Chai2022Selective,
title={Selective Data Acquisition in the Wild for Model Charging},
author={Cheng-Liang Chai and Jia-Bin Liu and N. Tang and Guoliang Li and Yu-Yu Luo},
journal={Proc. VLDB Endow.},
year={2022},
volume={15},
pages={1466-1478},
doi={10.14778/3523210.3523223}
}

@article{Tharwat2023SurveyAL,
title={A Survey on Active Learning: State-of-the-Art, Practical Challenges and Research Directions},
author={Alaa Tharwat and Wolfram Schenck},
journal={Mathematics},
year={2023},
doi={10.3390/math11040820}
}

@article{Li2023Label-Efficient,
title={Label-Efficient Learning in Agriculture: A Comprehensive Review},
author={Jiajia Li and Dong Chen and Xin-Da Qi and Zhao Li and Yan-Bo Huang and Daniel Morris and Xiaobo Tan},
journal={Comput. Electron. Agric.},
year={2023},
volume={215},
pages={108412},
doi={10.48550/arxiv.2305.14691}
}

\newpage
\appendix
\onecolumn

\section{Implementation Details}

\subsection{Hyperparameter Settings}
\label{appdx:hyperparameters}

We evaluate all models with at most 100 labels, no validation data is available, so all models use their default settings and all acquisition rules the values below, which are identical for every dataset and seed.

\textbf{Frozen TFMs.}
TabICL uses the publicly released v2 classification checkpoint (package \texttt{tabicl}~2.2.0) and TabPFN the v3 classification checkpoints, the binary one for binary and the multiclass one for multiclass tasks (package \texttt{tabpfn}~9.2.0, with the pre-training limits on the numbers of rows and features disabled). Both use an ensemble of eight members, their default preprocessing and inference settings, and the seed of the run as random state. No parameter of either model is updated in any experiment; the model is refitted, i.e., conditioned on the current labeled set, after every acquisition round.

\textbf{Protocol.}
For every seed $0,\dots,9$, a stratified 70/30 split into pool and test set; an initial context of one uniformly drawn pool row per class; batches of $b=10$ rows; budget $B=\min(100,\lfloor|\Pool|/2\rfloor)$; Cohen's $\kappa$ on the test split after every round. Ties in the win/tie/loss counts are differences of at most 0.005.

\textbf{Feature space.}
Coreset, Hybrid, Proxy-Hybrid and TANN-feat measure distances in the feature space standardized with the mean and standard deviation of the pool; missing values are set to zero after standardization.

\textbf{Tab-AICL rules.}
\emph{Margin} labels the $b$ pool rows with the largest margin uncertainty $1-(p_{(1)}-p_{(2)})$, where $p_{(1)}$ and $p_{(2)}$ are the two largest predicted class probabilities. \emph{Coreset} selects greedily the $b$ pool rows farthest (squared Euclidean distance) from the labeled set ($k$-center). \emph{Hybrid} restricts the unlabeled rows $\Pool_r=\Pool\setminus\Lab_r$ to the $\max(2b,\lfloor|\Pool_r|/2\rfloor)$ rows with the largest predictive entropy, runs $k$-means with $k=b$ on them (scikit-learn, automatic number of initializations, seed of the run) and labels the row nearest to each centroid. \emph{Proxy-Hybrid} fits a logistic regression with balanced class weights (at most 1{,}000 iterations) on the labeled rows, shortlists the $\min(|\Pool_r|,\max(200,\min(2000,\lfloor0.05|\Pool_r|\rfloor)))$ rows with the largest proxy entropy, keeps the $3b$ of them with the largest TFM entropy, and applies the $k$-means step of Hybrid.

\textbf{Target-aware rules.}
The targets are the inputs of the test split. \emph{TANN} chooses the $b$ targets with the largest margin uncertainty under the current context and assigns to each its nearest unlabeled pool row, without repetition. Distances are Euclidean in the TFM's row representation, recomputed in every round, after centering and scaling each row's representation across its dimensions and normalizing it to unit length; \emph{TANN-feat} uses the standardized feature space instead. \emph{Target-Random} assigns nearest pool rows to $b$ uniformly drawn targets and \emph{Target-Coverage} to $b$ targets chosen greedily as farthest from the labeled rows and the targets already chosen, both in the row representation.

\textbf{EPIC.}
In every round, $m_r=256$ targets are drawn uniformly without replacement and a shortlist of $s=64$ candidates is formed from the 21 unlabeled rows with the largest margin uncertainty, the nearest unlabeled rows to the most uncertain targets in the row representation (up to 42 candidates in total) and uniformly drawn unlabeled rows for the remainder. The hypothetical contexts $\Data_\Lab\cup\{(x,y)\}$ for all $x$ in the shortlist and $y\in\Y$ are evaluated in batched forward passes of 64 contexts, and the $b$ candidates with the largest EPIC score are labeled. With TabICL, the preprocessing statistics are fitted once on $\Data_\Lab$ and the shortlist with placeholder labels and shared among the hypothetical contexts; with TabPFN, every hypothetical context is preprocessed separately (\cref{appdx:tabpfn}).

\textbf{Baselines with other models.}
XGBoost (package \texttt{xgboost}~3.4.1, histogram tree method) and CatBoost (package \texttt{catboost}~1.2.10) use their library defaults and are refitted on the labeled rows in every round; Margin uses their predicted probabilities. LabelSpreading (scikit-learn~1.9.1) uses a $k$-nearest-neighbor kernel with seven neighbors and at most 100 iterations on the standardized features of the labeled rows, the pool and the test inputs.

\textbf{Label reduction.}
The budget is 10\% of the pool and the batch size 0.5\% of the pool (rounded up); EPIC labels at most 32 rows per round, so its larger batches are split into several rounds. $\kappa$ at 1\%, 2\%, 5\% and 10\% of the pool is interpolated linearly from the learning curve and divided by the $\kappa$ of the same backbone with the fully labeled pool as context on the same split.

\textbf{Context analysis.}
Local contexts consist of the $K\in\{25,50\}$ labeled rows nearest to a query in the row representation, completed so that every class is present, and of $K=100$ rows when the whole pool is labeled. Pseudo-labeled rows are pool rows that TabICL, conditioned on the acquired labels, predicts with probability at least 0.9; the per-query variants add the 50 such rows nearest to the query. SMOTE generates 500 synthetic rows by interpolating between an acquired row and one of its five nearest acquired rows of the same class in the standardized feature space, of which the 50 nearest to the query are added.

\textbf{Software and hardware.}
PyTorch~2.14.1; every run used one NVIDIA A40 GPU (48\,GB) and 16 CPU cores.

\subsection{Pseudo-Code}
\label{appdx:code}

\begin{algorithm}[H]
  \caption{EPIC: expected predictive information gain in context of TFMs}
  \label{alg:icepig}
  \begin{algorithmic}
    \STATE {\bfseries Input:} pool $\Pool$, targets $\Tgt$, frozen TFM $p_\theta$,
      budget $B$, batch size $b$, shortlist size $s$, target sample size $m_r$
    \STATE $\Lab\leftarrow$ one uniformly drawn row per class; query its labels
    \WHILE{$|\Lab|<B$}
      \STATE fit $p_\theta$ on $\Data_\Lab$ \hfill\COMMENT{no parameter update}
      \STATE $\Tgt_r\leftarrow$ uniform subsample of $\Tgt$ of size $m_r$
      \STATE $\Cand_r\leftarrow$ shortlist of $s$ unlabeled rows \hfill\COMMENT{\cref{sec:method-estimation}}
      \STATE fit $p_\theta$ once on $\Data_\Lab$ and $\Cand_r$ with placeholder labels
      \STATE compute $p_\theta(y^\star\mid x^\star,\Data_\Lab\cup\{(x,y)\})$ for all
        $x\in\Cand_r$, $y\in\Y$, $x^\star\in\Tgt_r$ \hfill\COMMENT{batched}
      \STATE compute $\mathrm{EPIC}(x;\Data_\Lab)$ for all $x\in\Cand_r$ \hfill\COMMENT{\cref{eq:icepig}}
      \STATE $A\leftarrow$ the $\min(b,B-|\Lab|)$ candidates with the largest EPIC score
      \STATE query the labels of $A$; $\Lab\leftarrow\Lab\cup A$
    \ENDWHILE
    \STATE {\bfseries Predict:} $p_\theta(\cdot\mid x^\star,\Data_\Lab)$ for all $x^\star\in\Tgt$
  \end{algorithmic}
\end{algorithm}

\section{Benchmark Datasets: Details}
\label{appdx:benchmarkDatasets}
\Cref{tab:datasets} summarizes the dataset statistics of the  twenty OpenML~\citep{vanschoren2014openml} classification datasets used in our evaluation. The setup encompasses the same setup as in Tab-AICL \cite{treerath2026tabaicl}.

\begin{table}[H]
  \centering
  \small
  \setlength{\tabcolsep}{4pt}
  \caption{Benchmark datasets (OpenML identifiers as in Tab-AICL). Rows after uniform
    subsampling to at most 10{,}000 (original size in parentheses), number of features
    $d$ and classes $C$, and the label budget $B$, which is 100 or half of the pool.}
  \label{tab:datasets}
  \begin{tabular}{lrrrcr}
    \toprule
    Dataset & OpenML & Rows & $d$ & $C$ & $B$ \\
    \midrule
    Iris & 61 & 150 & 4 & 3 & 52 \\
    Glass & 41 & 214 & 9 & 6 & 74 \\
    Ionosphere & 59 & 351 & 34 & 2 & 100 \\
    Balance-scale & 997 & 625 & 4 & 2 & 100 \\
    Vehicle & 994 & 846 & 18 & 2 & 100 \\
    Page-blocks & 1021 & 5{,}473 & 10 & 2 & 100 \\
    Parkinsons & 1488 & 195 & 22 & 2 & 68 \\
    Seeds & 1499 & 210 & 7 & 3 & 73 \\
    Bank-Marketing & 45065 & 10{,}000 (45{,}211) & 16 & 2 & 100 \\
    Adult & 45068 & 10{,}000 (48{,}842) & 14 & 2 & 100 \\
    CoverType & 1596 & 10{,}000 (581{,}012) & 54 & 7 & 100 \\
    KC1 & 1066 & 145 & 94 & 2 & 50 \\
    JM1 & 46979 & 10{,}000 (10{,}885) & 21 & 2 & 100 \\
    Blood-Transfusion & 46913 & 748 & 4 & 2 & 100 \\
    Diabetes & 46921 & 768 & 8 & 2 & 100 \\
    Tic-Tac-Toe & 137 & 10{,}000 (39{,}366) & 9 & 2 & 100 \\
    Credit-g & 46918 & 1{,}000 & 20 & 2 & 100 \\
    Steel-Plates & 46959 & 1{,}941 & 26 & 7 & 100 \\
    Phoneme & 44127 & 3{,}172 & 5 & 2 & 100 \\
    ILPD & 41943 & 583 & 10 & 2 & 100 \\
    \bottomrule
  \end{tabular}
\end{table}

\section{Labels vs. Context}
\label{appdx:context}

\paragraph{Labels or context?}
A natural alternative to acquiring better labels is to make better use of the labels one has by reshaping the context. For every dataset and seed, we took the labeled sets that Random, Hybrid, TANN and EPIC acquired up to the budget and evaluated them with restricted and with enlarged contexts (\cref{tab:context-suite}). 

\emph{Restricting} the context to the labeled rows nearest to each query in TabICL's representation lowers $\kappa$ for all four rules: by 0.04 to 0.05 with $K=25$ rows (worse on 15 to 17 of 20 datasets) and by about 0.02 with $K=50$. Even when the whole pool is labeled, local contexts of 100 rows remain below the global context (0.640 versus 0.661, worse on 13 datasets), which is in line with the observation that local contexts retain most, but not all, of the full-context accuracy and are motivated by inference cost~\citep{guta2026localized}. With at most 100 acquired labels, the global context is small, and localization only discards information that the model exploits.

\emph{Enlarging} the context with every pool row that TabICL predicts with probability at least 0.9, labeled with its prediction, yields a small and consistent gain of 0.009 to 0.015 (better on 11 to 13 datasets, worse on at most three; uncorrected $p\le0.015$ for every rule, significant after Holm correction for TANN). These pseudo-labels are 93 to 94\% correct on average, and the errors matter: adding the 50 pseudo-labeled rows nearest to each query has no effect, whereas the same rows with their true labels improve $\kappa$ by 0.020 to 0.034. Synthetic SMOTE neighbors lower $\kappa$ for all rules (worse on 13 to 15 datasets).

This provides evidence for two conclusions. First, the effect of reshaping the context is small compared with the effect of acquisition: the best manipulation that requires no further labels adds about 0.01, whereas EPIC improves on Random by 0.04 and on Hybrid by 0.03 with the same number of labels. Second, the ordering of the four rules at the budget (EPIC, TANN, Hybrid, Random) is the same under every context variant, so the comparison of \cref{sec:eval-how} does not depend on how the acquired labels are presented to the model, and confident pseudo-labeling is a complementary addition to, not a substitute for, the choice of labels.

\begin{table*}[t]
\centering
\small
\setlength{\tabcolsep}{5pt}
\caption{Reshaping the context of the acquired labels on the twenty Tab-AICL benchmarks (frozen TabICL, 10 seeds). First row: Cohen's $\kappa$ at the budget with the acquired labels as one global context. The acquired labels are those of the rerun in \cref{tab:transfer}, so the $\kappa$ of EPIC differs from \cref{tab:final} (0.578) because of GPU nondeterminism. Other rows: mean difference to that context over datasets and win/tie/loss (tie: $|\Delta|\le 0.005$); $^{*}$: two-sided Wilcoxon signed-rank test over datasets, $p<0.05$ after Holm correction over all comparisons of the table. Local contexts contain the $K$ labelled rows nearest to each query in TabICL's row representation. Augmented contexts contain all acquired labels plus either every pool row that TabICL labels with probability $\ge 0.9$, with its predicted label (confident pseudo-labels), or, per query, the 50 nearest such rows with predicted or true labels, or the 50 nearest SMOTE rows. $K=50$: 19 datasets (on KC1 the budget is 50 labels).}
\label{tab:context-suite}
\begin{tabular}{lcccc}
\toprule
Context & Random & Hybrid & TANN & EPIC \\
\midrule
Global context ($\kappa$) & 0.536 & 0.549 & 0.570 & 0.579 \\
\midrule
Local context, $K=25$ & $-0.043$$^{*}$\,{\scriptsize 3/0/17} & $-0.037$$^{*}$\,{\scriptsize 3/2/15} & $-0.040$$^{*}$\,{\scriptsize 2/2/16} & $-0.046$$^{*}$\,{\scriptsize 1/2/17} \\
Local context, $K=50$ & $-0.018$\,{\scriptsize 2/4/13} & $-0.019$$^{*}$\,{\scriptsize 2/4/13} & $-0.020$$^{*}$\,{\scriptsize 2/5/12} & $-0.013$\,{\scriptsize 2/4/13} \\
\addlinespace
+ confident pseudo-labels & $+0.013$\,{\scriptsize 12/7/1} & $+0.015$\,{\scriptsize 12/5/3} & $+0.010$$^{*}$\,{\scriptsize 13/7/0} & $+0.009$\,{\scriptsize 11/8/1} \\
+ pseudo-labelled neighbours & $-0.001$\,{\scriptsize 9/6/5} & $+0.005$\,{\scriptsize 10/5/5} & $+0.003$\,{\scriptsize 7/11/2} & $-0.000$\,{\scriptsize 3/13/4} \\
+ same neighbours, true labels (oracle) & $+0.026$$^{*}$\,{\scriptsize 16/2/2} & $+0.034$$^{*}$\,{\scriptsize 16/4/0} & $+0.022$$^{*}$\,{\scriptsize 15/4/1} & $+0.020$$^{*}$\,{\scriptsize 13/6/1} \\
+ synthetic neighbours (SMOTE) & $-0.014$\,{\scriptsize 5/2/13} & $-0.012$\,{\scriptsize 3/2/15} & $-0.016$\,{\scriptsize 4/1/15} & $-0.016$\,{\scriptsize 3/3/14} \\
\midrule
\multicolumn{5}{l}{\emph{All pool labels:} global context $\kappa=0.661$; local context, $K=100$: $-0.021$\,{\scriptsize 3/4/13}} \\
\bottomrule
\end{tabular}
\end{table*}

\section{Additional Results with TabICL}
\label{sec:eval-secondary}

\subsection{When: the value of acquisition criteria changes as the context grows}
\label{appdx:when_learningCurves}

\begin{figure}
    \centering
    \includegraphics[width=0.5\columnwidth]{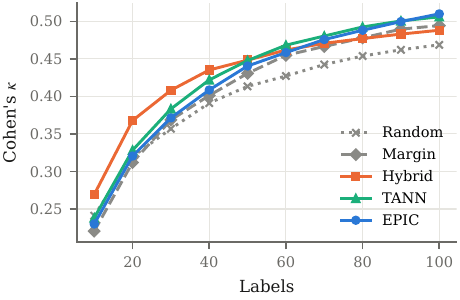}
    \caption{Learning curves on the fifteen datasets with a budget of 100 labels, with TabICL as the backbone. Cohen’s $\kappa$ after each acquisition round, averaged over datasets (each averaged over seeds)}
    \label{fig:learning_curves_mean}
\end{figure}

\textbf{Diversity leads in the cold start, prediction-oriented acquisition at the budget.}
\Cref{fig:learning_curves_mean} shows the learning curves on the fifteen datasets with a budget of 100 labels. We observe that in cold start Hybrid leads but converges subpar to the proposed acquisition rules of TANN and EPIC. 

\subsection{How: EPIC computes prediction-oriented information gain in context - Other baselines}
\label{appdx:how_baselines}
\textbf{Best models at the budget.~}
Against the baselines of other models (\cref{tab:baselines}), EPIC improves on XGBoost-Margin and LabelSpreading-Random on all twenty datasets, on CatBoost-Margin on 18, and on every rule run with TabPFN~v3 on 13 to 18 datasets ($p\le0.03$); with the same acquisition rule, TabICL attains a higher AULC than TabPFN for all five Tab-AICL rules (13 to 16 of 20 datasets, $p\le0.014$). The improvement of EPIC at the budget also holds for two secondary metrics that we record alongside $\kappa$, balanced accuracy (0.756 vs.\ 0.742 for Hybrid) and macro-averaged one-vs-rest ROC AUC (0.875 vs.\ 0.870).
\begin{table}[h]
\centering
\small
\setlength{\tabcolsep}{3pt}
\caption{Comparison with baselines using other models on 20 datasets (mean over datasets of the seed means; ranks among all listed methods). The last columns compare EPIC with each method in final $\kappa$ (W/T/L, two-sided Wilcoxon $p$). TabPFN rules use the frozen TabPFN v3; gradient-boosting and label spreading baselines are refitted every round.}
\label{tab:baselines}
\resizebox{0.5\textwidth}{!}{
\begin{tabular}{lccccc}
\toprule
& \multicolumn{2}{c}{AULC $\kappa$} & \multicolumn{2}{c}{Final $\kappa$} & EPIC vs. \\
\cmidrule(lr){2-3}\cmidrule(lr){4-5}
Method & Mean & Rank & Mean & Rank & W/T/L ($p$) \\
\midrule
EPIC & 0.479 & 3.85 & 0.578 & 3.35 & -- \\
TANN & 0.479 & 3.80 & 0.570 & 4.85 & 11/4/5 (0.133) \\
Hybrid & 0.482 & 3.55 & 0.549 & 6.42 & 15/1/4 (0.011) \\
Margin & 0.467 & 5.50 & 0.563 & 5.58 & 13/4/3 (0.024) \\
Random & 0.453 & 6.90 & 0.536 & 7.65 & 16/2/2 (0.000) \\
\midrule
TabPFN-Random & 0.419 & 9.40 & 0.518 & 8.88 & 18/0/2 (0.000) \\
TabPFN-Margin & 0.437 & 8.00 & 0.542 & 6.50 & 14/1/5 (0.004) \\
TabPFN-Coreset & 0.382 & 9.90 & 0.489 & 9.95 & 16/3/1 (0.000) \\
TabPFN-Hybrid & 0.456 & 6.05 & 0.534 & 7.67 & 16/1/3 (0.000) \\
TabPFN-Proxy-Hybrid & 0.435 & 7.90 & 0.551 & 5.92 & 14/2/4 (0.017) \\
TabPFN-TANN-feat & 0.444 & 7.35 & 0.553 & 5.72 & 13/1/6 (0.027) \\
XGBoost-Margin & 0.374 & 11.15 & 0.498 & 10.80 & 20/0/0 (0.000) \\
CatBoost-Margin & 0.426 & 9.50 & 0.525 & 8.75 & 18/1/1 (0.000) \\
LabelSpreading-Random & 0.351 & 12.15 & 0.424 & 12.95 & 20/0/0 (0.000) \\
\bottomrule
\end{tabular}
}
\end{table}

\textbf{Fewer labels for the same quality. }
\label{appdx:how_reduction}
\Cref{tab:reduction} summarizes the performance. 
On Page-blocks, 5\% of the pool (192 labels) suffices for 98\% of the full-context $\kappa$ and 10\% for practically all of it; with 10\% of the labels, EPIC retains 95\% on Adult, 91\% on Phoneme, 81--87\% on CoverType, Bank-Marketing, Tic-Tac-Toe and Steel-Plates, and 73\% on JM1, whose full-context $\kappa$ is only 0.22. 
Because $\kappa$ discounts chance agreement, it is a strict measure of retention; in accuracy, EPIC retains 93.5\% and 96.2\% of the full-context value with 5\% and 10\% of the labels (Random: 91.3\% and 93.9\%). 
Labeling a few percent of the pool thus recovers \textit{most}, but generally not all, of the full-data performance.

\begin{table}[h]
\centering
\small
\setlength{\tabcolsep}{3.5pt}
\caption{Label reduction on 8 datasets (pools of 1,358 to 7,000 rows): Cohen's $\kappa$ with a fraction of the pool labeled, relative to TabICL with the fully labeled pool (in \%, mean over datasets of the seed means; 100\% = full-context performance). In parentheses: datasets reaching at least 95\% of the full-context $\kappa$ out of those where the fraction is defined.}
\label{tab:reduction}
\begin{tabular}{lcccc}
\toprule
Rule & \multicolumn{4}{c}{Labelled fraction of the pool} \\
\cmidrule(lr){2-5}
 & 1\% & 2\% & 5\% & 10\% \\
\midrule
Random & 55.7 (0/8) & 62.6 (0/8) & 72.5 (0/8) & 78.9 (0/8) \\
Margin & 51.2 (0/8) & 59.0 (0/8) & 73.7 (1/8) & 83.6 (1/8) \\
Hybrid & \textbf{60.4 (0/8)} & \textbf{66.0 (0/8)} & 76.0 (0/8) & 82.0 (0/8) \\
TANN & 50.6 (0/8) & 61.9 (0/8) & 75.8 (1/8) & 83.5 (1/8) \\
EPIC & 50.9 (0/8) & 64.8 (0/8) & \textbf{80.3 (1/8)} & \textbf{87.0 (1/8)} \\
\bottomrule
\end{tabular}
\end{table}

\subsection{Cost and the information signal}
\label{appdx:cost}
EPIC takes 48\,s per run on the small and 145\,s on the large datasets at the 100-label budget, roughly an order of magnitude more than the other rules, and 2 to 30 minutes per run when labeling up to 10\% of a pool of up to 7{,}000 rows; its cost does not grow with the pool beyond the shortlist. As an instrument, the largest information gain of a round decreases steadily with the number of labels, from about $10^{-1}$ to about $2\cdot10^{-3}$ nats over the budget. Used as a stopping rule, i.e., stopping a run once this value falls below a threshold $\tau$, it improves only marginally on stopping every run at the same fraction of its budget, by at most 0.010 $\kappa$ (\cref{tab:stopping}).

\begin{table}[t]
\centering
\small
\caption{Stopping EPIC once the largest information gain of a round falls below $\tau$ (post hoc). Labels used as a fraction of the budget, $\kappa$ at the stopping point, and $\kappa$ when every run instead stops at the same fraction of its budget.}
\label{tab:stopping}
\begin{tabular}{ccccc}
\toprule
$\tau$ & Labels used & $\kappa$ (adaptive) & $\kappa$ (fixed fraction) & $\kappa$ (full budget) \\
\midrule
0.02 & 39.6\% & 0.493 & 0.483 & 0.578 \\
0.01 & 57.2\% & 0.526 & 0.526 & 0.578 \\
0.005 & 72.5\% & 0.554 & 0.550 & 0.578 \\
0.002 & 91.2\% & 0.574 & 0.569 & 0.578 \\
0.001 & 97.4\% & 0.577 & 0.576 & 0.578 \\
\bottomrule
\end{tabular}
\end{table}

\subsection{Analysis: whole trajectory}
\begin{figure}
    \centering
    \includegraphics[width=0.5\linewidth]{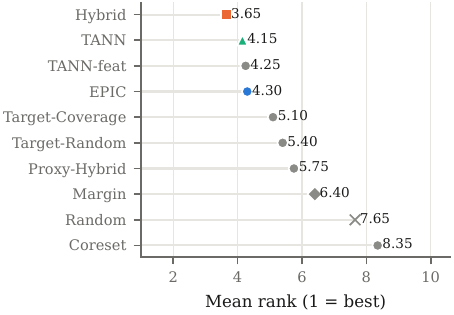}
    \caption{Mean rank of every acquisition rule over the twenty datasets (1 is best), for the AULC of Cohen’s $\kappa$, with TabICL as the backbone.}
    \label{fig:ranks_aucl}
\end{figure}
As shown in \Cref{fig:ranks_aucl}, we observe that the two primary metrics (AULC and the evaluation of $\kappa$ at the label budget as shown in \Cref{fig:ranks_final}) make the rules differently.
Over the whole trajectory, Hybrid, which combines uncertainty with $k$-means diversity, has the best mean AULC (0.482) and mean rank (3.65, \cref{tab:aulc}); EPIC (0.479, rank 4.30) and
TANN (0.479, rank 4.15) are statistically indistinguishable from it (EPIC vs.\
Hybrid: 7/2/11 datasets, $p=0.70$). However, in the learning curves, we see that with increasing label budget, TANN and EPIC win over Hybrid.

\begin{table*}[h]
\centering
\small
\setlength{\tabcolsep}{3pt}
\caption{AULC of Cohen's $\kappa$ (mean $\pm$ std over 10 seeds) on the Tab-AICL benchmarks with TabICL; best in bold, second underlined.}
\label{tab:aulc}
\resizebox{\textwidth}{!}{
\begin{tabular}{lcccccccccc}
\toprule
Dataset & \rotatebox{0}{Random} & \rotatebox{0}{Margin} & \rotatebox{0}{Coreset} & \rotatebox{0}{Hybrid} & \makecell{Proxy-\\Hybrid} & \makecell{Target-\\Random} & \makecell{Target-\\Coverage} & \rotatebox{0}{TANN-feat} & \rotatebox{0}{TANN} & \rotatebox{0}{EPIC} \\
\midrule
iris & 0.884\,{\scriptsize$\pm$0.051} & 0.877\,{\scriptsize$\pm$0.045} & 0.890\,{\scriptsize$\pm$0.039} & \textbf{0.897\,{\scriptsize$\pm$0.029}} & \underline{0.891\,{\scriptsize$\pm$0.037}} & 0.889\,{\scriptsize$\pm$0.043} & 0.887\,{\scriptsize$\pm$0.034} & 0.891\,{\scriptsize$\pm$0.039} & 0.885\,{\scriptsize$\pm$0.048} & 0.884\,{\scriptsize$\pm$0.043} \\
glass & 0.499\,{\scriptsize$\pm$0.037} & 0.516\,{\scriptsize$\pm$0.060} & 0.441\,{\scriptsize$\pm$0.051} & 0.524\,{\scriptsize$\pm$0.044} & 0.519\,{\scriptsize$\pm$0.037} & 0.540\,{\scriptsize$\pm$0.046} & 0.525\,{\scriptsize$\pm$0.058} & 0.551\,{\scriptsize$\pm$0.054} & \underline{0.554\,{\scriptsize$\pm$0.062}} & \textbf{0.586\,{\scriptsize$\pm$0.046}} \\
ionosphere & 0.734\,{\scriptsize$\pm$0.042} & 0.744\,{\scriptsize$\pm$0.042} & 0.525\,{\scriptsize$\pm$0.089} & \textbf{0.788\,{\scriptsize$\pm$0.034}} & 0.759\,{\scriptsize$\pm$0.045} & 0.759\,{\scriptsize$\pm$0.061} & 0.706\,{\scriptsize$\pm$0.077} & 0.730\,{\scriptsize$\pm$0.060} & \underline{0.760\,{\scriptsize$\pm$0.044}} & 0.719\,{\scriptsize$\pm$0.038} \\
balance-scale & 0.711\,{\scriptsize$\pm$0.042} & 0.770\,{\scriptsize$\pm$0.048} & 0.756\,{\scriptsize$\pm$0.029} & \textbf{0.780\,{\scriptsize$\pm$0.005}} & 0.769\,{\scriptsize$\pm$0.017} & 0.735\,{\scriptsize$\pm$0.048} & 0.753\,{\scriptsize$\pm$0.048} & 0.767\,{\scriptsize$\pm$0.031} & 0.764\,{\scriptsize$\pm$0.042} & \underline{0.772\,{\scriptsize$\pm$0.053}} \\
vehicle & 0.781\,{\scriptsize$\pm$0.037} & 0.769\,{\scriptsize$\pm$0.080} & 0.763\,{\scriptsize$\pm$0.045} & \textbf{0.843\,{\scriptsize$\pm$0.032}} & 0.770\,{\scriptsize$\pm$0.055} & 0.785\,{\scriptsize$\pm$0.035} & 0.758\,{\scriptsize$\pm$0.057} & 0.802\,{\scriptsize$\pm$0.038} & 0.796\,{\scriptsize$\pm$0.033} & \underline{0.809\,{\scriptsize$\pm$0.036}} \\
page-blocks & 0.640\,{\scriptsize$\pm$0.044} & 0.663\,{\scriptsize$\pm$0.043} & 0.485\,{\scriptsize$\pm$0.119} & \textbf{0.720\,{\scriptsize$\pm$0.024}} & 0.700\,{\scriptsize$\pm$0.031} & 0.636\,{\scriptsize$\pm$0.023} & 0.694\,{\scriptsize$\pm$0.061} & \underline{0.711\,{\scriptsize$\pm$0.029}} & 0.706\,{\scriptsize$\pm$0.049} & 0.666\,{\scriptsize$\pm$0.054} \\
parkinsons & 0.570\,{\scriptsize$\pm$0.124} & 0.605\,{\scriptsize$\pm$0.144} & 0.591\,{\scriptsize$\pm$0.125} & 0.636\,{\scriptsize$\pm$0.107} & \underline{0.662\,{\scriptsize$\pm$0.114}} & 0.626\,{\scriptsize$\pm$0.101} & \textbf{0.672\,{\scriptsize$\pm$0.116}} & 0.654\,{\scriptsize$\pm$0.080} & 0.626\,{\scriptsize$\pm$0.119} & 0.650\,{\scriptsize$\pm$0.104} \\
seeds & 0.883\,{\scriptsize$\pm$0.021} & 0.880\,{\scriptsize$\pm$0.020} & \textbf{0.890\,{\scriptsize$\pm$0.021}} & 0.884\,{\scriptsize$\pm$0.024} & 0.881\,{\scriptsize$\pm$0.023} & 0.880\,{\scriptsize$\pm$0.030} & \underline{0.888\,{\scriptsize$\pm$0.017}} & 0.881\,{\scriptsize$\pm$0.022} & 0.886\,{\scriptsize$\pm$0.022} & 0.877\,{\scriptsize$\pm$0.026} \\
bank-marketing & 0.191\,{\scriptsize$\pm$0.058} & \underline{0.255\,{\scriptsize$\pm$0.071}} & 0.222\,{\scriptsize$\pm$0.063} & 0.175\,{\scriptsize$\pm$0.072} & 0.237\,{\scriptsize$\pm$0.060} & \textbf{0.255\,{\scriptsize$\pm$0.062}} & 0.226\,{\scriptsize$\pm$0.055} & 0.167\,{\scriptsize$\pm$0.053} & 0.230\,{\scriptsize$\pm$0.057} & 0.203\,{\scriptsize$\pm$0.064} \\
adult & 0.362\,{\scriptsize$\pm$0.056} & 0.381\,{\scriptsize$\pm$0.054} & 0.343\,{\scriptsize$\pm$0.047} & \textbf{0.411\,{\scriptsize$\pm$0.035}} & 0.389\,{\scriptsize$\pm$0.035} & \underline{0.405\,{\scriptsize$\pm$0.045}} & 0.400\,{\scriptsize$\pm$0.029} & 0.390\,{\scriptsize$\pm$0.028} & 0.392\,{\scriptsize$\pm$0.057} & 0.399\,{\scriptsize$\pm$0.050} \\
covertype & 0.304\,{\scriptsize$\pm$0.023} & 0.331\,{\scriptsize$\pm$0.037} & 0.290\,{\scriptsize$\pm$0.013} & \underline{0.333\,{\scriptsize$\pm$0.026}} & 0.305\,{\scriptsize$\pm$0.049} & 0.321\,{\scriptsize$\pm$0.031} & 0.315\,{\scriptsize$\pm$0.034} & \textbf{0.335\,{\scriptsize$\pm$0.040}} & 0.327\,{\scriptsize$\pm$0.041} & 0.323\,{\scriptsize$\pm$0.035} \\
kc1 & 0.422\,{\scriptsize$\pm$0.124} & \underline{0.457\,{\scriptsize$\pm$0.086}} & 0.348\,{\scriptsize$\pm$0.089} & 0.398\,{\scriptsize$\pm$0.100} & 0.365\,{\scriptsize$\pm$0.092} & 0.383\,{\scriptsize$\pm$0.135} & 0.424\,{\scriptsize$\pm$0.062} & 0.453\,{\scriptsize$\pm$0.094} & 0.396\,{\scriptsize$\pm$0.103} & \textbf{0.462\,{\scriptsize$\pm$0.084}} \\
jm1 & \underline{0.145\,{\scriptsize$\pm$0.071}} & 0.087\,{\scriptsize$\pm$0.072} & 0.066\,{\scriptsize$\pm$0.080} & \textbf{0.145\,{\scriptsize$\pm$0.039}} & 0.088\,{\scriptsize$\pm$0.045} & 0.113\,{\scriptsize$\pm$0.060} & 0.099\,{\scriptsize$\pm$0.039} & 0.062\,{\scriptsize$\pm$0.067} & 0.081\,{\scriptsize$\pm$0.064} & 0.108\,{\scriptsize$\pm$0.052} \\
blood-transfusion & 0.158\,{\scriptsize$\pm$0.052} & 0.173\,{\scriptsize$\pm$0.062} & 0.201\,{\scriptsize$\pm$0.075} & \underline{0.207\,{\scriptsize$\pm$0.053}} & 0.160\,{\scriptsize$\pm$0.084} & 0.181\,{\scriptsize$\pm$0.052} & 0.174\,{\scriptsize$\pm$0.088} & 0.177\,{\scriptsize$\pm$0.088} & \textbf{0.215\,{\scriptsize$\pm$0.092}} & 0.182\,{\scriptsize$\pm$0.098} \\
diabetes & 0.337\,{\scriptsize$\pm$0.027} & 0.351\,{\scriptsize$\pm$0.062} & 0.336\,{\scriptsize$\pm$0.039} & 0.365\,{\scriptsize$\pm$0.066} & 0.338\,{\scriptsize$\pm$0.058} & 0.352\,{\scriptsize$\pm$0.059} & \underline{0.373\,{\scriptsize$\pm$0.041}} & \textbf{0.379\,{\scriptsize$\pm$0.040}} & 0.365\,{\scriptsize$\pm$0.043} & 0.369\,{\scriptsize$\pm$0.057} \\
tic-tac-toe & 0.190\,{\scriptsize$\pm$0.045} & 0.198\,{\scriptsize$\pm$0.065} & 0.143\,{\scriptsize$\pm$0.052} & 0.178\,{\scriptsize$\pm$0.050} & 0.204\,{\scriptsize$\pm$0.036} & 0.200\,{\scriptsize$\pm$0.042} & \textbf{0.222\,{\scriptsize$\pm$0.057}} & \underline{0.208\,{\scriptsize$\pm$0.071}} & 0.206\,{\scriptsize$\pm$0.063} & 0.207\,{\scriptsize$\pm$0.050} \\
credit-g & 0.183\,{\scriptsize$\pm$0.073} & 0.182\,{\scriptsize$\pm$0.053} & 0.171\,{\scriptsize$\pm$0.040} & \textbf{0.241\,{\scriptsize$\pm$0.036}} & 0.209\,{\scriptsize$\pm$0.069} & 0.201\,{\scriptsize$\pm$0.054} & 0.193\,{\scriptsize$\pm$0.050} & \underline{0.212\,{\scriptsize$\pm$0.060}} & 0.194\,{\scriptsize$\pm$0.028} & 0.195\,{\scriptsize$\pm$0.069} \\
steel-plates & 0.462\,{\scriptsize$\pm$0.027} & 0.472\,{\scriptsize$\pm$0.037} & 0.438\,{\scriptsize$\pm$0.046} & 0.502\,{\scriptsize$\pm$0.026} & 0.430\,{\scriptsize$\pm$0.050} & 0.473\,{\scriptsize$\pm$0.032} & 0.488\,{\scriptsize$\pm$0.029} & \textbf{0.512\,{\scriptsize$\pm$0.019}} & \underline{0.510\,{\scriptsize$\pm$0.028}} & 0.502\,{\scriptsize$\pm$0.038} \\
phoneme & 0.473\,{\scriptsize$\pm$0.033} & 0.482\,{\scriptsize$\pm$0.050} & 0.454\,{\scriptsize$\pm$0.031} & 0.469\,{\scriptsize$\pm$0.051} & 0.458\,{\scriptsize$\pm$0.124} & 0.469\,{\scriptsize$\pm$0.028} & 0.474\,{\scriptsize$\pm$0.047} & \underline{0.497\,{\scriptsize$\pm$0.065}} & \textbf{0.515\,{\scriptsize$\pm$0.053}} & 0.495\,{\scriptsize$\pm$0.049} \\
ilpd & 0.138\,{\scriptsize$\pm$0.050} & 0.150\,{\scriptsize$\pm$0.055} & 0.127\,{\scriptsize$\pm$0.045} & 0.153\,{\scriptsize$\pm$0.062} & 0.179\,{\scriptsize$\pm$0.079} & \textbf{0.180\,{\scriptsize$\pm$0.056}} & 0.163\,{\scriptsize$\pm$0.085} & 0.140\,{\scriptsize$\pm$0.057} & 0.168\,{\scriptsize$\pm$0.060} & \underline{0.180\,{\scriptsize$\pm$0.047}} \\
\midrule
Mean & 0.453 & 0.467 & 0.424 & 0.482 & 0.466 & 0.469 & 0.472 & 0.476 & 0.479 & 0.479 \\
Mean rank $\downarrow$ & 7.65 & 6.40 & 8.35 & \textbf{3.65} & 5.75 & 5.40 & 5.10 & 4.25 & 4.15 & 4.30 \\
\bottomrule
\end{tabular}
}
\end{table*}

\section{TabPFN as the Backbone}
\label{appdx:tabpfn}

All results in the main text use TabICL as the frozen TFM. To test whether the findings depend on this backbone, we repeated the complete study with TabPFN~v3~\citep{hollmann2025tabpfn} (binary and multiclass classification checkpoints, eight ensemble members) under the identical protocol: the same twenty datasets, splits, seeds, initial contexts, batch size and budgets, and all ten acquisition rules, including TANN, the two controls and EPIC. Only the backbone changes.

\textbf{Implementation.}
The target-aware rules require the TFM's row representation and EPIC additionally the model's predictions under hypothetical contexts. For TabPFN, the row representation is the transformer's output for the query rows, conditioned on the current context and averaged over the ensemble members, analogous to TabICL. For EPIC, every hypothetical context $\Data_\Lab\cup\{(x,y)\}$ is evaluated by an exact forward pass with its own preprocessing, batched over contexts, instead of sharing the preprocessing statistics among the hypothetical contexts as for TabICL (\cref{sec:method-estimation}). With TabPFN, EPIC takes 253\,s and 437\,s per run on the small and large datasets, compared with 48\,s and 145\,s with TabICL, which we attribute to the per-context preprocessing; TANN takes 13\,s and 22\,s.

\textbf{Overall agreement.}
\Cref{tab:tabpfn-vs-tabicl} juxtaposes every rule with both backbones. TabPFN attains a lower mean $\kappa$ than TabICL for every rule except Proxy-Hybrid at the budget (by 0.01 to 0.04), but the rules are ordered almost identically: the Spearman correlation of the rule means between the two backbones is 0.94 for the AULC and 0.89 for $\kappa$ at the budget. The only notable change concerns Proxy-Hybrid, which ranks considerably better with TabPFN (4.15 vs.\ 6.70 at the budget).

\begin{table}[t]
\centering
\small
\setlength{\tabcolsep}{3pt}
\caption{Every rule with TabICL and TabPFN as the frozen TFM on the 20 datasets: mean Cohen's $\kappa$ over datasets (mean rank in parentheses; best per column in bold). Spearman correlations of the rule means between the two TFMs: $\rho=0.94$ (AULC), $\rho=0.89$ (final).}
\label{tab:tabpfn-vs-tabicl}
\begin{tabular}{lcccc}
\toprule
& \multicolumn{2}{c}{AULC $\kappa$} & \multicolumn{2}{c}{Final $\kappa$} \\
\cmidrule(lr){2-3}\cmidrule(lr){4-5}
Rule & TabICL & TabPFN & TabICL & TabPFN \\
\midrule
Random & 0.453 (7.65) & 0.419 (7.10) & 0.536 (6.90) & 0.518 (7.03) \\
Margin & 0.467 (6.40) & 0.437 (5.90) & 0.563 (5.30) & 0.542 (5.15) \\
Coreset & 0.424 (8.35) & 0.382 (7.30) & 0.520 (8.45) & 0.489 (7.88) \\
Hybrid & \textbf{0.482 (3.65)} & \textbf{0.456 (3.75)} & 0.549 (5.88) & 0.534 (5.55) \\
Proxy-Hybrid & 0.466 (5.75) & 0.435 (5.70) & 0.550 (6.70) & 0.551 (4.15) \\
Target-Random & 0.469 (5.40) & 0.427 (6.60) & 0.556 (5.72) & 0.526 (7.03) \\
Target-Coverage & 0.472 (5.10) & 0.446 (4.80) & 0.564 (4.58) & 0.536 (6.00) \\
TANN-feat & 0.476 (4.25) & 0.444 (5.05) & 0.569 (3.77) & 0.553 (4.05) \\
TANN & \underline{0.479 (4.15)} & \underline{0.450 (4.10)} & \underline{0.570 (4.40)} & \underline{0.554 (4.22)} \\
EPIC & \underline{0.479 (4.30)} & 0.447 (4.70) & \textbf{0.578 (3.30)} & \textbf{0.558 (3.95)} \\
\bottomrule
\end{tabular}
\end{table}

\begin{table*}[h]
\centering
\small
\setlength{\tabcolsep}{3pt}
\caption{Cohen's $\kappa$ at the label budget with TabPFN (mean $\pm$ std over 10 seeds); best in bold, second underlined.}
\label{tab:tabpfn-final}
\resizebox{\textwidth}{!}{
\begin{tabular}{lcccccccccc}
\toprule
Dataset & \rotatebox{0}{Random} & \rotatebox{0}{Margin} & \rotatebox{0}{Coreset} & \rotatebox{0}{Hybrid} & \makecell{Proxy-Hybrid} & \makecell{Target-\\Random} & \makecell{Target-\\Coverage} & \rotatebox{0}{TANN-feat} & \rotatebox{0}{TANN} & \rotatebox{0}{EPIC} \\
\midrule
iris & \textbf{0.957\,{\scriptsize$\pm$0.030}} & 0.950\,{\scriptsize$\pm$0.045} & 0.943\,{\scriptsize$\pm$0.042} & \underline{0.957\,{\scriptsize$\pm$0.052}} & 0.957\,{\scriptsize$\pm$0.033} & 0.933\,{\scriptsize$\pm$0.039} & 0.950\,{\scriptsize$\pm$0.040} & 0.957\,{\scriptsize$\pm$0.042} & 0.957\,{\scriptsize$\pm$0.033} & 0.940\,{\scriptsize$\pm$0.042} \\
glass & 0.609\,{\scriptsize$\pm$0.054} & 0.616\,{\scriptsize$\pm$0.052} & 0.546\,{\scriptsize$\pm$0.090} & 0.593\,{\scriptsize$\pm$0.035} & 0.601\,{\scriptsize$\pm$0.085} & 0.641\,{\scriptsize$\pm$0.065} & 0.590\,{\scriptsize$\pm$0.074} & 0.665\,{\scriptsize$\pm$0.076} & \underline{0.667\,{\scriptsize$\pm$0.061}} & \textbf{0.680\,{\scriptsize$\pm$0.066}} \\
ionosphere & 0.869\,{\scriptsize$\pm$0.045} & \textbf{0.894\,{\scriptsize$\pm$0.040}} & 0.619\,{\scriptsize$\pm$0.102} & 0.877\,{\scriptsize$\pm$0.041} & \underline{0.890\,{\scriptsize$\pm$0.048}} & 0.869\,{\scriptsize$\pm$0.037} & 0.871\,{\scriptsize$\pm$0.025} & 0.828\,{\scriptsize$\pm$0.051} & 0.886\,{\scriptsize$\pm$0.040} & 0.874\,{\scriptsize$\pm$0.043} \\
balance-scale & 0.847\,{\scriptsize$\pm$0.033} & 0.938\,{\scriptsize$\pm$0.018} & 0.882\,{\scriptsize$\pm$0.028} & 0.919\,{\scriptsize$\pm$0.021} & \textbf{0.941\,{\scriptsize$\pm$0.016}} & 0.879\,{\scriptsize$\pm$0.025} & 0.868\,{\scriptsize$\pm$0.032} & 0.900\,{\scriptsize$\pm$0.041} & 0.928\,{\scriptsize$\pm$0.021} & \underline{0.940\,{\scriptsize$\pm$0.014}} \\
vehicle & 0.946\,{\scriptsize$\pm$0.019} & \underline{0.995\,{\scriptsize$\pm$0.007}} & 0.927\,{\scriptsize$\pm$0.041} & 0.969\,{\scriptsize$\pm$0.019} & 0.994\,{\scriptsize$\pm$0.007} & 0.950\,{\scriptsize$\pm$0.029} & 0.959\,{\scriptsize$\pm$0.022} & 0.989\,{\scriptsize$\pm$0.007} & \textbf{0.996\,{\scriptsize$\pm$0.005}} & 0.993\,{\scriptsize$\pm$0.008} \\
page-blocks & 0.673\,{\scriptsize$\pm$0.061} & 0.774\,{\scriptsize$\pm$0.056} & 0.714\,{\scriptsize$\pm$0.073} & 0.755\,{\scriptsize$\pm$0.054} & 0.782\,{\scriptsize$\pm$0.076} & 0.697\,{\scriptsize$\pm$0.065} & 0.797\,{\scriptsize$\pm$0.037} & \textbf{0.833\,{\scriptsize$\pm$0.019}} & \underline{0.830\,{\scriptsize$\pm$0.021}} & 0.815\,{\scriptsize$\pm$0.022} \\
parkinsons & 0.726\,{\scriptsize$\pm$0.110} & 0.744\,{\scriptsize$\pm$0.085} & 0.671\,{\scriptsize$\pm$0.120} & 0.774\,{\scriptsize$\pm$0.097} & \textbf{0.791\,{\scriptsize$\pm$0.100}} & 0.741\,{\scriptsize$\pm$0.088} & 0.778\,{\scriptsize$\pm$0.092} & \underline{0.789\,{\scriptsize$\pm$0.118}} & 0.778\,{\scriptsize$\pm$0.093} & 0.780\,{\scriptsize$\pm$0.099} \\
seeds & 0.948\,{\scriptsize$\pm$0.028} & 0.943\,{\scriptsize$\pm$0.032} & 0.933\,{\scriptsize$\pm$0.037} & \textbf{0.952\,{\scriptsize$\pm$0.032}} & 0.940\,{\scriptsize$\pm$0.031} & 0.933\,{\scriptsize$\pm$0.037} & 0.943\,{\scriptsize$\pm$0.027} & \underline{0.950\,{\scriptsize$\pm$0.027}} & 0.948\,{\scriptsize$\pm$0.032} & 0.938\,{\scriptsize$\pm$0.034} \\
bank-marketing & 0.190\,{\scriptsize$\pm$0.102} & 0.203\,{\scriptsize$\pm$0.107} & \underline{0.254\,{\scriptsize$\pm$0.121}} & 0.198\,{\scriptsize$\pm$0.124} & \textbf{0.324\,{\scriptsize$\pm$0.059}} & 0.240\,{\scriptsize$\pm$0.130} & 0.195\,{\scriptsize$\pm$0.059} & 0.236\,{\scriptsize$\pm$0.091} & 0.232\,{\scriptsize$\pm$0.080} & 0.234\,{\scriptsize$\pm$0.076} \\
adult & 0.420\,{\scriptsize$\pm$0.076} & 0.477\,{\scriptsize$\pm$0.074} & 0.371\,{\scriptsize$\pm$0.089} & 0.464\,{\scriptsize$\pm$0.060} & \textbf{0.500\,{\scriptsize$\pm$0.031}} & 0.430\,{\scriptsize$\pm$0.104} & 0.475\,{\scriptsize$\pm$0.063} & 0.460\,{\scriptsize$\pm$0.071} & 0.450\,{\scriptsize$\pm$0.153} & \underline{0.490\,{\scriptsize$\pm$0.049}} \\
covertype & 0.447\,{\scriptsize$\pm$0.028} & \underline{0.465\,{\scriptsize$\pm$0.014}} & 0.461\,{\scriptsize$\pm$0.034} & 0.463\,{\scriptsize$\pm$0.019} & 0.433\,{\scriptsize$\pm$0.035} & 0.444\,{\scriptsize$\pm$0.030} & 0.448\,{\scriptsize$\pm$0.025} & 0.462\,{\scriptsize$\pm$0.024} & \textbf{0.476\,{\scriptsize$\pm$0.015}} & 0.458\,{\scriptsize$\pm$0.020} \\
kc1 & 0.491\,{\scriptsize$\pm$0.128} & 0.417\,{\scriptsize$\pm$0.127} & 0.456\,{\scriptsize$\pm$0.112} & 0.425\,{\scriptsize$\pm$0.133} & 0.439\,{\scriptsize$\pm$0.127} & 0.457\,{\scriptsize$\pm$0.108} & 0.466\,{\scriptsize$\pm$0.144} & \underline{0.494\,{\scriptsize$\pm$0.097}} & 0.490\,{\scriptsize$\pm$0.151} & \textbf{0.495\,{\scriptsize$\pm$0.097}} \\
jm1 & \textbf{0.117\,{\scriptsize$\pm$0.062}} & 0.079\,{\scriptsize$\pm$0.066} & 0.099\,{\scriptsize$\pm$0.071} & \underline{0.112\,{\scriptsize$\pm$0.040}} & 0.088\,{\scriptsize$\pm$0.072} & 0.084\,{\scriptsize$\pm$0.071} & 0.080\,{\scriptsize$\pm$0.029} & 0.067\,{\scriptsize$\pm$0.049} & 0.066\,{\scriptsize$\pm$0.052} & 0.069\,{\scriptsize$\pm$0.060} \\
blood-transfusion & 0.094\,{\scriptsize$\pm$0.088} & \textbf{0.271\,{\scriptsize$\pm$0.050}} & 0.105\,{\scriptsize$\pm$0.077} & 0.199\,{\scriptsize$\pm$0.085} & 0.191\,{\scriptsize$\pm$0.092} & 0.169\,{\scriptsize$\pm$0.091} & 0.091\,{\scriptsize$\pm$0.079} & \underline{0.212\,{\scriptsize$\pm$0.097}} & 0.190\,{\scriptsize$\pm$0.074} & 0.153\,{\scriptsize$\pm$0.069} \\
diabetes & 0.419\,{\scriptsize$\pm$0.070} & 0.411\,{\scriptsize$\pm$0.053} & 0.433\,{\scriptsize$\pm$0.037} & 0.422\,{\scriptsize$\pm$0.044} & \textbf{0.455\,{\scriptsize$\pm$0.044}} & 0.409\,{\scriptsize$\pm$0.062} & 0.421\,{\scriptsize$\pm$0.073} & \underline{0.451\,{\scriptsize$\pm$0.074}} & 0.423\,{\scriptsize$\pm$0.053} & 0.443\,{\scriptsize$\pm$0.065} \\
tic-tac-toe & 0.202\,{\scriptsize$\pm$0.075} & 0.182\,{\scriptsize$\pm$0.084} & 0.123\,{\scriptsize$\pm$0.083} & 0.191\,{\scriptsize$\pm$0.048} & 0.220\,{\scriptsize$\pm$0.053} & 0.154\,{\scriptsize$\pm$0.125} & \textbf{0.270\,{\scriptsize$\pm$0.033}} & 0.235\,{\scriptsize$\pm$0.095} & \underline{0.266\,{\scriptsize$\pm$0.100}} & 0.260\,{\scriptsize$\pm$0.076} \\
credit-g & 0.195\,{\scriptsize$\pm$0.128} & \textbf{0.238\,{\scriptsize$\pm$0.113}} & 0.181\,{\scriptsize$\pm$0.092} & 0.190\,{\scriptsize$\pm$0.063} & \underline{0.228\,{\scriptsize$\pm$0.069}} & 0.163\,{\scriptsize$\pm$0.122} & 0.188\,{\scriptsize$\pm$0.093} & 0.220\,{\scriptsize$\pm$0.093} & 0.216\,{\scriptsize$\pm$0.113} & 0.227\,{\scriptsize$\pm$0.101} \\
steel-plates & 0.551\,{\scriptsize$\pm$0.036} & 0.585\,{\scriptsize$\pm$0.021} & 0.457\,{\scriptsize$\pm$0.095} & 0.604\,{\scriptsize$\pm$0.034} & 0.565\,{\scriptsize$\pm$0.028} & 0.588\,{\scriptsize$\pm$0.036} & 0.594\,{\scriptsize$\pm$0.038} & \textbf{0.621\,{\scriptsize$\pm$0.032}} & 0.577\,{\scriptsize$\pm$0.039} & \underline{0.621\,{\scriptsize$\pm$0.025}} \\
phoneme & 0.554\,{\scriptsize$\pm$0.038} & 0.568\,{\scriptsize$\pm$0.062} & 0.507\,{\scriptsize$\pm$0.032} & 0.537\,{\scriptsize$\pm$0.047} & 0.563\,{\scriptsize$\pm$0.056} & 0.591\,{\scriptsize$\pm$0.040} & 0.563\,{\scriptsize$\pm$0.048} & \underline{0.632\,{\scriptsize$\pm$0.027}} & 0.608\,{\scriptsize$\pm$0.047} & \textbf{0.633\,{\scriptsize$\pm$0.021}} \\
ilpd & 0.101\,{\scriptsize$\pm$0.074} & 0.089\,{\scriptsize$\pm$0.075} & 0.090\,{\scriptsize$\pm$0.100} & 0.084\,{\scriptsize$\pm$0.094} & 0.119\,{\scriptsize$\pm$0.104} & \underline{0.139\,{\scriptsize$\pm$0.087}} & \textbf{0.173\,{\scriptsize$\pm$0.137}} & 0.063\,{\scriptsize$\pm$0.077} & 0.093\,{\scriptsize$\pm$0.099} & 0.121\,{\scriptsize$\pm$0.099} \\
\midrule
Mean & 0.518 & 0.542 & 0.489 & 0.534 & 0.551 & 0.526 & 0.536 & 0.553 & 0.554 & 0.558 \\
Mean rank $\downarrow$ & 7.03 & 5.15 & 7.88 & 5.55 & 4.15 & 7.03 & 6.00 & \underline{4.05} & 4.22 & \textbf{3.95} \\
\bottomrule
\end{tabular}
}
\end{table*}

\begin{table*}[t]
\centering
\small
\setlength{\tabcolsep}{3pt}
\caption{AULC of Cohen's $\kappa$ (mean $\pm$ std over 10 seeds) on the Tab-AICL benchmarks with TabPFN; best in bold, second underlined.}
\label{tab:tabpfn-aulc}
\resizebox{\textwidth}{!}{
\begin{tabular}{lcccccccccc}
\toprule
Dataset & \rotatebox{0}{Random} & \rotatebox{0}{Margin} & \rotatebox{0}{Coreset} & \rotatebox{0}{Hybrid} & \makecell{Proxy-\\Hybrid} & \makecell{Target-\\Random} & \makecell{Target-\\Coverage} & \rotatebox{0}{TANN-feat} & \rotatebox{0}{TANN} & \rotatebox{0}{EPIC} \\
\midrule
iris & 0.902\,{\scriptsize$\pm$0.045} & 0.902\,{\scriptsize$\pm$0.030} & \textbf{0.916\,{\scriptsize$\pm$0.041}} & 0.903\,{\scriptsize$\pm$0.040} & \underline{0.909\,{\scriptsize$\pm$0.034}} & 0.889\,{\scriptsize$\pm$0.034} & 0.900\,{\scriptsize$\pm$0.035} & 0.908\,{\scriptsize$\pm$0.036} & 0.907\,{\scriptsize$\pm$0.033} & 0.896\,{\scriptsize$\pm$0.038} \\
glass & 0.471\,{\scriptsize$\pm$0.039} & 0.478\,{\scriptsize$\pm$0.048} & 0.385\,{\scriptsize$\pm$0.041} & 0.472\,{\scriptsize$\pm$0.035} & 0.449\,{\scriptsize$\pm$0.048} & 0.515\,{\scriptsize$\pm$0.053} & 0.493\,{\scriptsize$\pm$0.060} & 0.529\,{\scriptsize$\pm$0.063} & \underline{0.540\,{\scriptsize$\pm$0.055}} & \textbf{0.551\,{\scriptsize$\pm$0.044}} \\
ionosphere & 0.702\,{\scriptsize$\pm$0.072} & 0.727\,{\scriptsize$\pm$0.034} & 0.346\,{\scriptsize$\pm$0.108} & \textbf{0.784\,{\scriptsize$\pm$0.036}} & \underline{0.771\,{\scriptsize$\pm$0.042}} & 0.706\,{\scriptsize$\pm$0.085} & 0.759\,{\scriptsize$\pm$0.038} & 0.705\,{\scriptsize$\pm$0.051} & 0.744\,{\scriptsize$\pm$0.057} & 0.651\,{\scriptsize$\pm$0.056} \\
balance-scale & 0.695\,{\scriptsize$\pm$0.052} & 0.703\,{\scriptsize$\pm$0.053} & 0.737\,{\scriptsize$\pm$0.025} & \textbf{0.752\,{\scriptsize$\pm$0.024}} & 0.714\,{\scriptsize$\pm$0.051} & 0.705\,{\scriptsize$\pm$0.039} & 0.691\,{\scriptsize$\pm$0.069} & 0.714\,{\scriptsize$\pm$0.049} & 0.718\,{\scriptsize$\pm$0.036} & \underline{0.739\,{\scriptsize$\pm$0.030}} \\
vehicle & 0.753\,{\scriptsize$\pm$0.049} & 0.790\,{\scriptsize$\pm$0.045} & 0.719\,{\scriptsize$\pm$0.048} & \textbf{0.822\,{\scriptsize$\pm$0.026}} & \underline{0.802\,{\scriptsize$\pm$0.039}} & 0.766\,{\scriptsize$\pm$0.059} & 0.756\,{\scriptsize$\pm$0.058} & 0.791\,{\scriptsize$\pm$0.038} & 0.796\,{\scriptsize$\pm$0.036} & 0.794\,{\scriptsize$\pm$0.062} \\
page-blocks & 0.576\,{\scriptsize$\pm$0.086} & 0.630\,{\scriptsize$\pm$0.060} & 0.447\,{\scriptsize$\pm$0.083} & \underline{0.689\,{\scriptsize$\pm$0.041}} & 0.645\,{\scriptsize$\pm$0.063} & 0.588\,{\scriptsize$\pm$0.051} & 0.678\,{\scriptsize$\pm$0.046} & 0.667\,{\scriptsize$\pm$0.059} & \textbf{0.693\,{\scriptsize$\pm$0.034}} & 0.635\,{\scriptsize$\pm$0.061} \\
parkinsons & 0.517\,{\scriptsize$\pm$0.125} & 0.580\,{\scriptsize$\pm$0.119} & 0.528\,{\scriptsize$\pm$0.109} & 0.571\,{\scriptsize$\pm$0.083} & 0.573\,{\scriptsize$\pm$0.095} & 0.589\,{\scriptsize$\pm$0.078} & 0.587\,{\scriptsize$\pm$0.128} & 0.581\,{\scriptsize$\pm$0.119} & \textbf{0.623\,{\scriptsize$\pm$0.130}} & \underline{0.606\,{\scriptsize$\pm$0.102}} \\
seeds & 0.892\,{\scriptsize$\pm$0.027} & 0.888\,{\scriptsize$\pm$0.023} & 0.890\,{\scriptsize$\pm$0.026} & \underline{0.901\,{\scriptsize$\pm$0.028}} & 0.880\,{\scriptsize$\pm$0.032} & 0.884\,{\scriptsize$\pm$0.034} & 0.897\,{\scriptsize$\pm$0.024} & 0.894\,{\scriptsize$\pm$0.023} & \textbf{0.902\,{\scriptsize$\pm$0.026}} & 0.877\,{\scriptsize$\pm$0.046} \\
bank-marketing & 0.120\,{\scriptsize$\pm$0.077} & 0.133\,{\scriptsize$\pm$0.078} & \underline{0.191\,{\scriptsize$\pm$0.094}} & 0.161\,{\scriptsize$\pm$0.112} & \textbf{0.229\,{\scriptsize$\pm$0.048}} & 0.128\,{\scriptsize$\pm$0.090} & 0.138\,{\scriptsize$\pm$0.046} & 0.144\,{\scriptsize$\pm$0.057} & 0.144\,{\scriptsize$\pm$0.062} & 0.136\,{\scriptsize$\pm$0.083} \\
adult & 0.296\,{\scriptsize$\pm$0.064} & \textbf{0.384\,{\scriptsize$\pm$0.065}} & 0.285\,{\scriptsize$\pm$0.061} & 0.369\,{\scriptsize$\pm$0.060} & 0.350\,{\scriptsize$\pm$0.045} & 0.286\,{\scriptsize$\pm$0.100} & \underline{0.376\,{\scriptsize$\pm$0.060}} & 0.363\,{\scriptsize$\pm$0.062} & 0.360\,{\scriptsize$\pm$0.124} & 0.363\,{\scriptsize$\pm$0.038} \\
covertype & 0.357\,{\scriptsize$\pm$0.021} & 0.374\,{\scriptsize$\pm$0.033} & 0.349\,{\scriptsize$\pm$0.030} & 0.370\,{\scriptsize$\pm$0.038} & 0.331\,{\scriptsize$\pm$0.034} & 0.343\,{\scriptsize$\pm$0.038} & 0.378\,{\scriptsize$\pm$0.030} & \underline{0.386\,{\scriptsize$\pm$0.027}} & \textbf{0.396\,{\scriptsize$\pm$0.023}} & 0.371\,{\scriptsize$\pm$0.027} \\
kc1 & \underline{0.389\,{\scriptsize$\pm$0.139}} & 0.331\,{\scriptsize$\pm$0.092} & 0.306\,{\scriptsize$\pm$0.077} & 0.349\,{\scriptsize$\pm$0.097} & 0.339\,{\scriptsize$\pm$0.084} & 0.381\,{\scriptsize$\pm$0.107} & 0.383\,{\scriptsize$\pm$0.087} & 0.358\,{\scriptsize$\pm$0.070} & 0.382\,{\scriptsize$\pm$0.113} & \textbf{0.405\,{\scriptsize$\pm$0.099}} \\
jm1 & \underline{0.102\,{\scriptsize$\pm$0.063}} & 0.061\,{\scriptsize$\pm$0.057} & 0.073\,{\scriptsize$\pm$0.049} & \textbf{0.111\,{\scriptsize$\pm$0.046}} & 0.069\,{\scriptsize$\pm$0.058} & 0.080\,{\scriptsize$\pm$0.061} & 0.068\,{\scriptsize$\pm$0.035} & 0.054\,{\scriptsize$\pm$0.045} & 0.042\,{\scriptsize$\pm$0.026} & 0.062\,{\scriptsize$\pm$0.056} \\
blood-transfusion & 0.078\,{\scriptsize$\pm$0.064} & \underline{0.160\,{\scriptsize$\pm$0.062}} & \textbf{0.163\,{\scriptsize$\pm$0.059}} & 0.158\,{\scriptsize$\pm$0.070} & 0.074\,{\scriptsize$\pm$0.025} & 0.121\,{\scriptsize$\pm$0.055} & 0.074\,{\scriptsize$\pm$0.041} & 0.119\,{\scriptsize$\pm$0.052} & 0.116\,{\scriptsize$\pm$0.060} & 0.114\,{\scriptsize$\pm$0.027} \\
diabetes & 0.345\,{\scriptsize$\pm$0.058} & 0.338\,{\scriptsize$\pm$0.056} & 0.314\,{\scriptsize$\pm$0.050} & \underline{0.376\,{\scriptsize$\pm$0.024}} & \textbf{0.395\,{\scriptsize$\pm$0.046}} & 0.351\,{\scriptsize$\pm$0.070} & 0.345\,{\scriptsize$\pm$0.110} & 0.328\,{\scriptsize$\pm$0.075} & 0.369\,{\scriptsize$\pm$0.040} & 0.360\,{\scriptsize$\pm$0.061} \\
tic-tac-toe & 0.080\,{\scriptsize$\pm$0.033} & 0.109\,{\scriptsize$\pm$0.043} & 0.060\,{\scriptsize$\pm$0.047} & 0.141\,{\scriptsize$\pm$0.049} & \underline{0.168\,{\scriptsize$\pm$0.056}} & 0.074\,{\scriptsize$\pm$0.055} & \textbf{0.178\,{\scriptsize$\pm$0.050}} & 0.149\,{\scriptsize$\pm$0.098} & 0.164\,{\scriptsize$\pm$0.098} & 0.139\,{\scriptsize$\pm$0.062} \\
credit-g & 0.117\,{\scriptsize$\pm$0.067} & 0.131\,{\scriptsize$\pm$0.091} & 0.143\,{\scriptsize$\pm$0.029} & \textbf{0.172\,{\scriptsize$\pm$0.072}} & 0.120\,{\scriptsize$\pm$0.072} & 0.098\,{\scriptsize$\pm$0.057} & 0.099\,{\scriptsize$\pm$0.063} & 0.114\,{\scriptsize$\pm$0.059} & 0.124\,{\scriptsize$\pm$0.090} & \underline{0.145\,{\scriptsize$\pm$0.068}} \\
steel-plates & 0.450\,{\scriptsize$\pm$0.030} & 0.463\,{\scriptsize$\pm$0.030} & 0.340\,{\scriptsize$\pm$0.066} & \underline{0.487\,{\scriptsize$\pm$0.024}} & 0.333\,{\scriptsize$\pm$0.055} & 0.463\,{\scriptsize$\pm$0.040} & 0.476\,{\scriptsize$\pm$0.035} & \textbf{0.499\,{\scriptsize$\pm$0.031}} & 0.449\,{\scriptsize$\pm$0.049} & 0.474\,{\scriptsize$\pm$0.025} \\
phoneme & 0.462\,{\scriptsize$\pm$0.041} & 0.472\,{\scriptsize$\pm$0.043} & 0.405\,{\scriptsize$\pm$0.036} & 0.461\,{\scriptsize$\pm$0.051} & 0.456\,{\scriptsize$\pm$0.052} & 0.461\,{\scriptsize$\pm$0.073} & 0.482\,{\scriptsize$\pm$0.046} & \textbf{0.513\,{\scriptsize$\pm$0.039}} & 0.464\,{\scriptsize$\pm$0.065} & \underline{0.504\,{\scriptsize$\pm$0.035}} \\
ilpd & 0.085\,{\scriptsize$\pm$0.047} & 0.095\,{\scriptsize$\pm$0.047} & 0.039\,{\scriptsize$\pm$0.039} & 0.070\,{\scriptsize$\pm$0.041} & 0.101\,{\scriptsize$\pm$0.097} & 0.113\,{\scriptsize$\pm$0.070} & \textbf{0.151\,{\scriptsize$\pm$0.117}} & 0.066\,{\scriptsize$\pm$0.041} & 0.071\,{\scriptsize$\pm$0.063} & \underline{0.114\,{\scriptsize$\pm$0.061}} \\
\midrule
Mean & 0.419 & 0.437 & 0.382 & 0.456 & 0.435 & 0.427 & 0.446 & 0.444 & 0.450 & 0.447 \\
Mean rank $\downarrow$ & 7.10 & 5.90 & 7.30 & \textbf{3.75} & 5.70 & 6.60 & 4.80 & 5.05 & \underline{4.10} & 4.70 \\
\bottomrule
\end{tabular}
}
\end{table*}

\begin{figure}[t]
  \centering
  \begin{subfigure}[t]{0.49\linewidth}
    \centering
    \includegraphics[width=\linewidth]{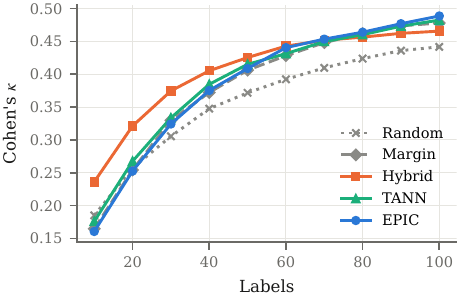}
    \caption{Mean Cohen's $\kappa$}
  \end{subfigure}\hfill
  \begin{subfigure}[t]{0.49\linewidth}
    \centering
    \includegraphics[width=\linewidth]{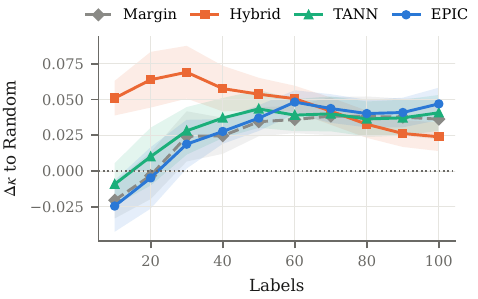}
    \caption{Difference to Random ($\pm1$ s.e.)}
  \end{subfigure}
  \caption{TabPFN as the backbone: learning curves on the fifteen datasets with a budget of 100 labels, averaged over datasets (each averaged over ten seeds). Hybrid leads in the cold start; EPIC and TANN overtake it after about 70 and 80 labels.}
  \label{fig:tabpfn-curves}
\end{figure}

\begin{figure}[t]
  \centering
  \begin{subfigure}[t]{0.49\linewidth}
    \centering
    \includegraphics[width=\linewidth]{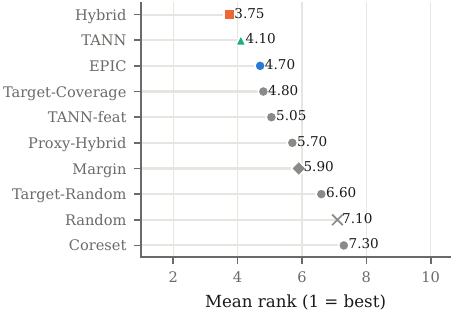}
    \caption{AULC of $\kappa$}
  \end{subfigure}\hfill
  \begin{subfigure}[t]{0.49\linewidth}
    \centering
    \includegraphics[width=\linewidth]{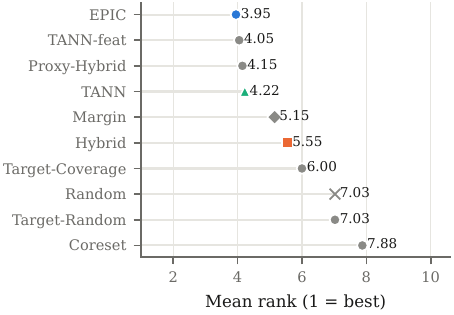}
    \caption{$\kappa$ at the budget}
  \end{subfigure}
  \caption{TabPFN as the backbone: mean rank of every acquisition rule over the twenty datasets (1 is best).}
  \label{fig:tabpfn-ranks}
\end{figure}

\begin{figure}[t]
  \centering
  \includegraphics[width=0.5\linewidth]{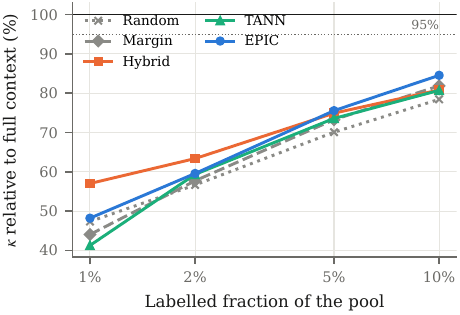}
  \caption{TabPFN as the backbone: Cohen's $\kappa$ with a fraction of the pool labeled, relative to TabPFN with the fully labeled pool, on the eight datasets with pools of 1{,}358 to 7{,}000 rows (mean over datasets of the means over ten seeds). The dotted line marks 95\% of the full-context performance.}
  \label{fig:tabpfn-reduction}
\end{figure}

\textbf{When.}
The transition from diversity- to prediction-oriented acquisition recurs (\cref{fig:tabpfn-curves}). Hybrid leads during the cold start, by up to 0.069 $\kappa$ over Random at 30 labels, and its advantage shrinks afterwards; EPIC overtakes it at about 70 labels and TANN at about 80, somewhat later than with TabICL. Accordingly, Hybrid has the best AULC and mean rank over the trajectory (0.456, rank 3.75; \cref{fig:tabpfn-ranks}, \cref{tab:tabpfn-aulc}), whereas EPIC attains the highest $\kappa$ and the best mean rank at the budget (0.558, rank 3.95; \cref{tab:tabpfn-final}) and improves on Hybrid on 14 of 20 datasets ($+0.024$, $p=0.017$; $p=0.086$ after Holm correction over nine comparisons), mirroring the TabICL result ($p=0.011$ and $0.064$). In the label-reduction study (\cref{fig:tabpfn-reduction}, \cref{tab:tabpfn-reduction}), Hybrid retains the most full-context $\kappa$ with 1\% and 2\% of the pool labeled (57.0\% and 63.4\%) and EPIC with 5\% and 10\% (75.5\% and 84.5\%; ahead of Hybrid on 7 of 8 datasets at both fractions, $p=0.20$ and $p=0.15$, and of Random on all eight at 10\%, $p=0.008$). The crossover is thus the same as with TabICL, although the individual differences are smaller.

\begin{table}[h]
\centering
\small
\setlength{\tabcolsep}{3.5pt}
\caption{TabPFN: label reduction on 8 datasets (pools of 1,358 to 7,000 rows): Cohen's $\kappa$ with a fraction of the pool labelled, relative to TabPFN with the fully labelled pool (in \%, mean over datasets of the seed means; 100\% = full-context performance). In parentheses: datasets reaching at least 95\% of the full-context $\kappa$ out of those where the fraction is defined.}
\label{tab:tabpfn-reduction}
\begin{tabular}{lcccc}
\toprule
Rule & \multicolumn{4}{c}{Labelled fraction of the pool} \\
\cmidrule(lr){2-5}
 & 1\% & 2\% & 5\% & 10\% \\
\midrule
Random & 47.3 (0/8) & 56.6 (0/8) & 70.1 (0/8) & 78.4 (0/8) \\
Margin & 44.0 (0/8) & 57.7 (0/8) & 73.3 (1/8) & 81.9 (2/8) \\
Hybrid & \textbf{57.0 (0/8)} & \textbf{63.4 (0/8)} & 74.9 (0/8) & 80.8 (1/8) \\
TANN & 41.3 (0/8) & 59.3 (0/8) & 73.6 (1/8) & 80.7 (2/8) \\
EPIC & 48.2 (0/8) & 59.6 (0/8) & \textbf{75.5 (1/8)} & \textbf{84.5 (2/8)} \\
\bottomrule
\end{tabular}
\end{table}

\textbf{Where.}
With TabPFN, access to the target inputs alone does not improve on random labeling: Target-Random is level with Random (AULC 10/2/8 datasets, $p=0.39$; at the budget 10/3/7, $p=0.29$), and Target-Coverage improves only the AULC (12/5/3, $+0.026$, $p=0.015$). Choosing the targets by the model's uncertainty or by information gain, however, matters even more than with TabICL (\cref{tab:tabpfn-pairwise}): at the budget, TANN improves on Target-Random on 16 datasets ($+0.028$, $p=0.002$, significant after Holm correction) and on Target-Coverage on 12 ($+0.018$, $p=0.030$), and EPIC improves on Target-Random on 14 ($+0.033$, $p=0.001$) and on Target-Coverage on 13 ($+0.022$, $p=0.007$), both significant after Holm correction. As with TabICL, TANN and EPIC are statistically indistinguishable (AULC 9/3/8, $p=0.93$; at the budget 7/5/8, $p=0.45$), as are TANN in the row representation and in the feature space (at the budget 6/8/6, $p=0.78$).
\begin{table}[h]
\centering
\small
\setlength{\tabcolsep}{3pt}
\caption{TabPFN: pairwise comparison across the 20 datasets: wins/ties/losses of the row method against each rule (tie: $|\Delta|\le0.005$), mean difference $\Delta$ of the dataset means, and the $p$-value of a two-sided Wilcoxon signed-rank test ($p<0.05$ in bold).}
\label{tab:tabpfn-pairwise}
\begin{tabular}{llcccccc}
\toprule
& & \multicolumn{3}{c}{AULC $\kappa$} & \multicolumn{3}{c}{Final $\kappa$} \\
\cmidrule(lr){3-5}\cmidrule(lr){6-8}
Method & vs. & W/T/L & $\Delta$ & $p$ & W/T/L & $\Delta$ & $p$ \\
\midrule
EPIC & Random & 16/0/4 & +0.027 & \textbf{0.004} & 15/2/3 & +0.040 & \textbf{0.001} \\
 & Margin & 10/5/5 & +0.010 & 0.123 & 11/3/6 & +0.016 & 0.064 \\
 & Coreset & 13/2/5 & +0.065 & \textbf{0.003} & 15/3/2 & +0.070 & \textbf{0.000} \\
 & Hybrid & 5/2/13 & -0.009 & 0.330 & 14/1/5 & +0.024 & \textbf{0.017} \\
 & Proxy-Hybrid & 11/1/8 & +0.012 & 0.245 & 7/5/8 & +0.007 & 0.729 \\
 & Target-Random & 15/1/4 & +0.020 & \textbf{0.006} & 14/2/4 & +0.033 & \textbf{0.001} \\
 & Target-Coverage & 9/3/8 & +0.001 & 0.648 & 13/3/4 & +0.022 & \textbf{0.007} \\
 & TANN-feat & 8/2/10 & +0.003 & 0.756 & 7/7/6 & +0.005 & 0.546 \\
 & TANN & 8/3/9 & -0.003 & 0.927 & 8/5/7 & +0.004 & 0.452 \\
\midrule
TANN & Random & 14/3/3 & +0.031 & \textbf{0.002} & 14/4/2 & +0.036 & \textbf{0.002} \\
 & Margin & 13/0/7 & +0.013 & 0.133 & 10/3/7 & +0.012 & 0.202 \\
 & Coreset & 14/0/6 & +0.068 & \textbf{0.004} & 16/1/3 & +0.065 & \textbf{0.000} \\
 & Hybrid & 5/5/10 & -0.006 & 0.430 & 12/4/4 & +0.020 & \textbf{0.040} \\
 & Proxy-Hybrid & 10/4/6 & +0.015 & 0.189 & 8/4/8 & +0.003 & 0.809 \\
 & Target-Random & 14/3/3 & +0.023 & \textbf{0.012} & 16/0/4 & +0.028 & \textbf{0.002} \\
 & Target-Coverage & 11/2/7 & +0.005 & 0.349 & 12/4/4 & +0.018 & \textbf{0.030} \\
 & TANN-feat & 11/6/3 & +0.006 & 0.097 & 6/8/6 & +0.001 & 0.778 \\
 & EPIC & 9/3/8 & +0.003 & 0.927 & 7/5/8 & -0.004 & 0.452 \\
\bottomrule
\end{tabular}
\end{table}

\begin{table}[h]
\centering
\small
\caption{Label transfer: final $\kappa$ of models trained on the labels acquired by each TabPFN rule (mean over datasets and seeds). The diagonal reference rows give the same model with its own margin-based acquisition.}
\label{tab:tabpfn-transfer}
\begin{tabular}{lccc}
\toprule
Labels acquired by & TabPFN & CatBoost & XGBoost \\
\midrule
Random & 0.518 & 0.500 & 0.487 \\
Margin & 0.542 & 0.526 & 0.507 \\
Hybrid & 0.534 & 0.520 & 0.501 \\
TANN & 0.554 & 0.532 & 0.519 \\
EPIC & 0.558 & 0.541 & 0.530 \\
\midrule
CatBoost-Margin (own labels) & -- & 0.525 & -- \\
XGBoost-Margin (own labels) & -- & -- & 0.498 \\
\bottomrule
\end{tabular}
\end{table}

\textbf{How.}
EPIC attains the highest mean $\kappa$ at the budget and improves on Random (15/2/3, $+0.040$), Coreset (15/3/2, $+0.070$) and both controls, all significant after Holm correction, and on Margin on 11 datasets ($+0.016$, $p=0.064$); Proxy-Hybrid and TANN are statistically level with it (\cref{tab:tabpfn-pairwise}). The labels acquired by EPIC with TabPFN again transfer to tree-based models (\cref{tab:tabpfn-transfer}): XGBoost and CatBoost trained on them reach a final $\kappa$ of 0.530 and 0.541, compared with 0.498 and 0.525 with their own margin-based acquisition (17/1/2 datasets, $p<0.001$, and 13/4/3, $p=0.012$), and the ordering Random $<$ Hybrid $<$ Margin $<$ TANN $<$ EPIC of the transferred labels is the same for both tree models.

Therefore, all three findings hold with TabPFN as the backbone: diversity leads in the cold start and prediction-oriented acquisition at the budget, choosing the targets to be served is essential, and EPIC and TANN yield the best models at the budget with labels that transfer to other models. The effect sizes and significance levels are of the same order as with TabICL.

\clearpage

\section{Ablation of EPIC's Approximations}
\label{appdx:epig-ablation}

EPIC approximates the information gain over all targets and all unlabeled rows in two ways (\cref{sec:method-estimation}): it averages over $m_r=\min(m,256)$ targets drawn per round instead of all $m=|\Tgt|$ targets, and it scores only a shortlist of $s$ candidates instead of every unlabeled row. We vary one of the two at a time with TabICL as the backbone, keeping the other at its default ($s=64$, $m_r=256$) and the protocol unchanged (batch size $b=10$, ten seeds). The shortlist size is varied on the eight datasets with the smallest pools (101 to 437 pool rows), on which scoring every unlabeled row is affordable; the number of sampled targets is varied on all twenty datasets (44 to 3{,}000 targets).

\begin{figure}[t]
  \centering
  \includegraphics[width=\linewidth]{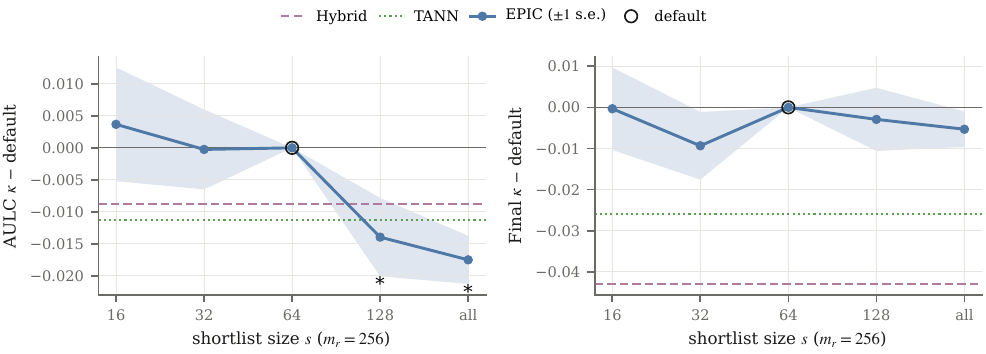}
  \caption{Ablation of the shortlist size $s$ of EPIC ($m_r=256$, TabICL) on the eight datasets with the smallest pools: difference of AULC $\kappa$ (left) and final $\kappa$ (right) to the default $s=64$ (circled), averaged over datasets ($\pm1$ standard error); ``all'' scores every unlabeled row. The horizontal lines mark Hybrid and TANN; $*$: two-sided Wilcoxon $p<0.05$ against the default.}
  \label{fig:epig-ablation-s}
\end{figure}

\textbf{Shortlist size.}
Final $\kappa$ is insensitive to the shortlist size: for $s\in\{16,32,128\}$ and for scoring every unlabeled row, it stays within 0.01 of the default without any significant difference, and every configuration remains well above TANN and Hybrid (\cref{fig:epig-ablation-s}, \cref{tab:epig-ablation-s}). Over the trajectory, however, larger shortlists are worse: with $s=128$ the AULC decreases by 0.014 (0/4/4 datasets, $p=0.04$) and with every unlabeled row scored by 0.017 (0/1/7, $p=0.008$), below the level of Hybrid and TANN, whereas $s=16$ and $s=32$ match the default. The shortlist is thus not only a reduction of cost. A plausible explanation is its composition: two thirds of it consist of rows with uncertain predictions and of rows near uncertain targets, and the remaining third is drawn at random, so it restricts the greedy top-$b$ selection to a diverse set of candidates; without this restriction, the $b$ highest-scoring rows of a round tend to carry overlapping information (\cref{sec:method-estimation}), which matters most in the cold start. In addition, with TabICL the shortlist rows enter the fit of the preprocessing statistics, so a larger shortlist also changes the statistics shared among the hypothetical contexts. The cost grows linearly with $s$, from 15\,s per run for $s=16$ to 108\,s when every unlabeled row is scored.

\begin{figure}[t]
  \centering
  \includegraphics[width=\linewidth]{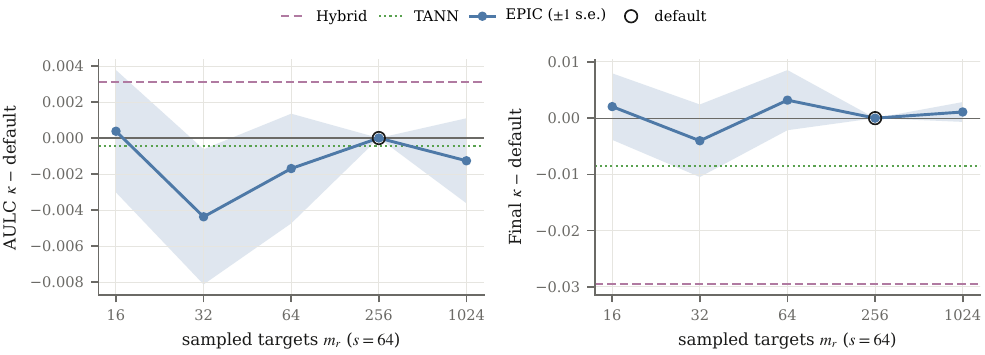}
  \caption{Ablation of the number of sampled targets $m_r$ of EPIC ($s=64$, TabICL) on the twenty datasets: difference of AULC $\kappa$ (left) and final $\kappa$ (right) to the default $m_r=256$ (circled), averaged over datasets ($\pm1$ standard error). On datasets with at most 256 targets, the default uses every target and $m_r=1024$ coincides with it. The horizontal lines mark Hybrid and TANN.}
  \label{fig:epig-ablation-m}
\end{figure}

\textbf{Number of sampled targets.}
The number of sampled targets has no measurable effect on the twenty datasets (\cref{fig:epig-ablation-m}, \cref{tab:epig-ablation-m}): for $m_r\in\{16,32,64,1024\}$, the mean final $\kappa$ and AULC stay within 0.005 of the default $m_r=256$, and no difference is significant ($p\ge0.08$); all configurations remain above Hybrid and TANN at the budget. This also holds on the eight datasets with more than 256 targets (583 to 3{,}000), on which the default uses a genuine subsample: averaging over four times as many targets ($m_r=1024$) changes final $\kappa$ by $+0.004$ ($p=0.46$) and AULC by $-0.004$ ($p=0.64$), and averaging over only $m_r=16$ by $+0.003$ ($p=1.00$) and $+0.006$ ($p=0.31$). On datasets with at most 256 targets, the default uses every target, so $m_r=1024$ coincides with it there. The cost, in contrast, grows with $m_r$; averaged over all twenty datasets, it ranges from 49\,s per run for $m_r=16$ to 87\,s for the default and 157\,s for $m_r=1024$.

Therefore, both approximations are benign at the budget: a few dozen sampled targets suffice to rank the candidates, and smaller shortlists are cheaper at equal quality, whereas larger ones reduce the quality over the trajectory. The defaults are therefore conservative, and EPIC could be run at a fraction of the reported cost.

\begin{table}[t]
\centering
\small
\setlength{\tabcolsep}{3pt}
\caption{Ablation of the shortlist size $s$ at $m_r=256$ sampled targets of EPIC with TabICL on 8 datasets (101--437 pool rows, 44--188 targets; default in bold). Mean Cohen\textquotesingle s $\kappa$ over datasets; $\Delta$: mean difference to the default (wins/ties/losses over datasets, tie $|\Delta|\le0.005$, two-sided Wilcoxon $p$). Time: mean seconds per acquisition run over the datasets of the table. With $s=\text{all}$, every unlabeled row is scored.}
\label{tab:epig-ablation-s}
\begin{tabular}{lclclr}
\toprule
& \multicolumn{2}{c}{Final $\kappa$} & \multicolumn{2}{c}{AULC $\kappa$} & \\
\cmidrule(lr){2-3}\cmidrule(lr){4-5}
$s$ & mean & $\Delta$ & mean & $\Delta$ & Time \\
\midrule
16 & 0.742 & $\pm0.000$ (4/1/3, 1.00) & 0.645 & $+0.004$ (6/0/2, 0.64) & 15 \\
32 & 0.733 & $-0.009$ (1/4/3, 0.56) & 0.641 & $\pm0.000$ (2/3/3, 0.84) & 24 \\
64 & \textbf{0.743} & -- & \textbf{0.641} & -- & 41 \\
128 & 0.740 & $-0.003$ (2/2/4, 0.46) & 0.627 & $-0.014$ (0/4/4, 0.04) & 65 \\
all & 0.737 & $-0.005$ (2/3/3, 0.25) & 0.624 & $-0.017$ (0/1/7, 0.01) & 108 \\
\midrule
Hybrid & 0.700 & $-0.043$ & 0.632 & $-0.009$ & -- \\
TANN & 0.717 & $-0.026$ & 0.630 & $-0.011$ & -- \\
\bottomrule
\end{tabular}
\end{table}

\begin{table}[t]
\centering
\small
\setlength{\tabcolsep}{3pt}
\caption{Ablation of the number of sampled targets $m_r$ at $s=64$ shortlisted candidates of EPIC with TabICL on 20 datasets (101--7{,}000 pool rows, 44--3{,}000 targets; default in bold). Mean Cohen\textquotesingle s $\kappa$ over datasets; $\Delta$: mean difference to the default (wins/ties/losses over datasets, tie $|\Delta|\le0.005$, two-sided Wilcoxon $p$). Time: mean seconds per acquisition run over the datasets of the table. Where a dataset has at most $\min(m_r,256)$ targets, the configuration uses every target, as the default does, and coincides with the default run.}
\label{tab:epig-ablation-m}
\begin{tabular}{lclclr}
\toprule
& \multicolumn{2}{c}{Final $\kappa$} & \multicolumn{2}{c}{AULC $\kappa$} & \\
\cmidrule(lr){2-3}\cmidrule(lr){4-5}
$m_r$ & mean & $\Delta$ & mean & $\Delta$ & Time \\
\midrule
16 & 0.580 & $+0.002$ (7/2/11, 0.84) & 0.480 & $\pm0.000$ (7/6/7, 0.96) & 49 \\
32 & 0.574 & $-0.004$ (6/4/10, 0.26) & 0.475 & $-0.004$ (3/5/12, 0.08) & 52 \\
64 & 0.581 & $+0.003$ (4/10/6, 0.47) & 0.478 & $-0.002$ (5/8/7, 0.53) & 58 \\
256 & \textbf{0.578} & -- & \textbf{0.479} & -- & 87 \\
1024 & 0.579 & $+0.001$ (4/12/4, 0.59) & 0.478 & $-0.001$ (4/11/5, 0.59) & 157 \\
\midrule
Hybrid & 0.549 & $-0.030$ & 0.482 & $+0.003$ & -- \\
TANN & 0.570 & $-0.008$ & 0.479 & $\pm0.000$ & -- \\
\bottomrule
\end{tabular}
\end{table}

\end{document}